\documentclass[10pt,journal,compsoc]{IEEEtran}

\ifCLASSINFOpdf
\else
\fi

\usepackage{psfrag,amsbsy,graphics,float}
\usepackage{verbatim} 
\usepackage{url}
\usepackage{cite}
\usepackage[numbers]{natbib}
\usepackage{wrapfig}
\usepackage{algorithm}
\usepackage[noend]{algorithmic}
\usepackage{amsmath}
\usepackage{amssymb}
\usepackage{amsthm}
\usepackage{bbm}
\usepackage{enumerate}
\usepackage{graphicx}
\usepackage{ifthen}
\usepackage{struktex}
\usepackage{hyperref}
\usepackage{color}
\usepackage{nicefrac} 
\usepackage{multirow}
\usepackage{hhline}
\usepackage{cleveref}
\usepackage{tabularx}
\usepackage{rotating}
\usepackage{caption}
\usepackage{subcaption}
\usepackage{rotating}
\usepackage{listings}
\usepackage{xcolor}
\usepackage{xtab}
\usepackage{diagbox}

\lstdefinestyle{Python}{
    language        = Python,
    basicstyle      = \ttfamily,
    keywordstyle    = \color{blue},
    keywordstyle    = [2] \color{teal}, % just to check that it works
    stringstyle     = \color{green},
    commentstyle    = \color{red}\ttfamily
}

\usepackage{booktabs}
\usepackage{makecell}
\usepackage{longtable}
\usepackage[nolist]{acronym}

\usepackage{adjustbox}

\usepackage{soul}

\usepackage[normalem]{ulem}

\usepackage{newfloat}
\DeclareFloatingEnvironment[name={Supplementary Figure},fileext=lsf,listname={List of Supplementary Figures}]{suppfigure}

\DeclareFloatingEnvironment[name={Supplementary Tables},fileext=lsf,listname={List of Supplementary Tables}]{supptable}

\newcommand{\ie}{i.\,e.}
\newcommand{\eg}{e.\,g.}
\newcommand{\wrt}{w.\,r.\,t. }

\newcommand{\ds}{\mbox{\textsc{Deep Spectrum }}}
\newcommand{\dsns}{\mbox{\textsc{Deep Spectrum}}}

\newcommand{\opensmile}{\mbox{\textsc{openSMILE }}}

\newcommand{\vggf}{\textsc{VGGface 2}}

\newcommand{\revision}[1] {{\color{black} #1}}

\begin{document}
%BS: added ``First''
\title{Towards Multimodal Prediction of Spontaneous Humor: A Novel Dataset and First Results}

\author{Lukas~Christ$^{1}$,
        Shahin~Amiriparian$^{2}$,
        Alexander~Kathan$^{1}$,
        Niklas~M\"uller$^{3}$,
        Andreas K\"onig$^{3}$,\\
        and~Bj\"orn~W.\ Schuller$^{1,2,4}$,~\IEEEmembership{Fellow,~IEEE}% <-this % stops a spacePh
        \thanks{$^{1}$Lukas Christ, Alexander Kathan and Bj\"orn W. Schuller are with the Chair of Embedded Intelligence for Health Care and Wellbeing, University of Augsburg, Germany. {\tt\small \{lukas1.christ,  alexander.kathan, bjoern.schuller\}@uni-a.de}}
\thanks{$^{2}$Shahin Amiriparian and Bj\"orn W. Schuller are with the Chair of Health Informatics, Klinikum rechts der Isar, TUM, Germany. {\tt\small \{shahin.amiriparian,  schuller\}@tum.de}}

\thanks{$^{3}$Niklas M\"uller and Andreas K\"onig are with the Chair of Strategic Management, Innovation, and Entrepreneurship, University of Passau, Germany. 
{\tt\small\{niklas.mueller, andreas.koenig\}@uni-passau.de}
}
\thanks{$^{4}$Bj\"{o}rn Schuller is also with GLAM -- the Group on Language, Audio, \& Music, Imperial College London, UK.}}
\markboth{Transactions on Affective Computing,~Vol.~xx, No.~x, x~2024}{}%
\IEEEtitleabstractindextext{%
\begin{abstract}
 
Humor is a substantial element of human social behavior, affect, and cognition. Its automatic understanding can facilitate a more naturalistic human-AI interaction. Current methods of humor detection have been exclusively based on staged data, making them inadequate for `real-world' applications. We contribute to addressing this deficiency by introducing the novel Passau-Spontaneous Football Coach Humor (Passau-SFCH) dataset, comprising about 11 hours of recordings. The Passau-SFCH dataset is annotated for the presence of humor and its dimensions (sentiment and direction) as proposed in Martin's Humor Style Questionnaire. We conduct a series of experiments employing pretrained Transformers, convolutional neural networks, and expert-designed features. The performance of each modality (text, audio, video) for spontaneous humor recognition is analyzed and their complementarity is investigated. Our findings suggest that for the automatic analysis of humor and its sentiment, facial expressions are most promising, while humor direction can be best modeled via text-based features. 
%The results highlight the individuality and contextuality of humor usage and style. 
Further, we experiment with different multimodal approaches to humor recognition, including decision-level fusion and MulT, a multimodal Transformer approach. In this context, we propose a novel multimodal architecture that yields the best overall results.
Finally, we make our code publicly available at \href{https://www.github.com/lc0197/passau-sfch}{https://www.github.com/lc0197/passau-sfch}. The Passau-SFCH dataset is available upon request.
\end{abstract}

}

\maketitle

% not be used in maketitle, but will appear (\ie, to be "transported")
\IEEEdisplaynontitleabstractindextext

% \fi
\IEEEpeerreviewmaketitle

\IEEEraisesectionheading{\section{Introduction}\label{sec:introduction}}


\noindent \IEEEPARstart{H}{umor} can be defined
as a communicative expression that establishes surprising or incongruent relationships or meaning and is intended to amuse~\cite{berger1976anatomy, cooper2005just}. Humor is an essential element of human communication and is characterized by its multifaceted nature~\cite{ruch1998temperament}. Whether in the form of a simple punchline or humorous non-verbal behavior within a conversation, humor can be used in a variety of ways and can have strong effects on interlocutors' states of mind and interpersonal perceptions, making it a powerful form of human interaction~\cite{avolio1999funny,cooper2018leader}. In fact, humor impacts not only individuals but also group behavior and group dynamics. For example, certain types of humor, as used by leaders, have been shown to improve team performance in companies~\cite{wijewardena2017using}. More generally, humor can foster social exchange, relexation~\cite{cooper2018leader}, and creativity~\cite{wijewardena2017using}.

Scientific interest in humor can be traced back to at least the ancient Greeks and has led to a plethora of theories on the nature of humor~\cite{larkin2017overview}. As of today, numerous disciplines, including sociology~\cite{kuipers2008sociology}, psychology~\cite{martin2006psychology,goldstein2013psychology}, philosophy~\cite{morreall2014humor}, neuroscience~\cite{vrticka2013neural}, and management~\cite{cooper2005just} investigate humor from different perspectives.  In computer science and, especially, affective computing, the automatic measurement of humor has attracted increasing scholarly attention in recent years, \eg~\cite{weller2019humor,pramanick2022multimodal,hasan2021humor}. In particular, humor has been identified as important in human-computer interaction~\cite{weber2018shape, ritschel2020multimodal}. Because of the ubiquity of humor in our everyday interpersonal interactions, its automatic and multimodal measurement is highly relevant~\cite{avolio1999funny}.

Humor can be expressed in a variety of ways and modalities. An utterance may be perceived as humorous due to its semantic content (text), accompanying gestures or facial expressions (visual signals), variations in voice (audio signal) or a combination thereof. \citet{gironzetti2017prosodic} provides an extensive overview of empirical studies on the multimodality of humor.
Automated \ac{ML}-based humor recognition can exploit this multimodality by taking the textual, visual, and acoustic modalities into account, \eg~\cite{hasan2021humor}.

%\sout{In addition, lightweight ML frameworks such as \textsc{DeepSpectrumLite}~\cite{amiriparian2022deepspectrumlite} enable an on-device humor recognition~\cite{amiriparian2022deepspectrumlite}.}

However, previously presented multimodal \ac{ML} approaches are based on datasets in which humor is used in a scripted context, such as data collections from TED Talks~\cite{hasan2019ur} or TV series~\cite{bertero2016deep,wu2021mumor}. To the best of our knowledge, there is no database allowing for an automated detection of \emph{spontaneous} humor. This poses a difficulty as spontaneous \emph{in-the-wild} humor may substantially differ from planned humor in a scripted scenario. Moreover, existing datasets typically only come with a simple, binary label scheme indicating the presence of humor. 

%To address 
As a starting point to address these shortcomings, we introduce the novel \ac{Passau-SFCH} dataset which is based on press conferences -- and hence, non-scripted communication -- of professional football coaches from the German \emph{Bundesliga} 
%BS: added:
(the national premier soccer league). While press conferences are clearly not a fully \emph{in-the-wild} scenario, the dataset is intended to provide a first step towards humor recognition in more realistic, spontaneous communication situations.
\ac{Passau-SFCH} includes annotations according to the widely used \ac{HSQ} proposed by~\citet{martin2003individual}.
We compare the \ac{Passau-SFCH} dataset to other humor detection databases and provide deeper insights into the data.
 
Furthermore, we extract an extensive set of features for the audio, video, and text modalities. Based on that, we conduct a series of \ac{ML} experiments on automatically analysing the %\sout{spontaneous} 
use of humor in \ac{Passau-SFCH}.

\section{Related Work}
Early approaches to automatic humor recognition focused on textual data only. We give a brief overview of them in~\Cref{ssec:rw_text}. Recently, multimodal approaches, discussed in~\Cref{ssec:rw_multi}, have attracted increasing interest. 

\subsection{Textual Humor Recognition}\label{ssec:rw_text}
humor recognition has become a vibrant subfield of \ac{NLP}. 
The first attempts at text-based humor recognition can be traced back to 2004 when~\citet{taylor2004computationally} investigated wordplay in jokes utilizing N-grams. In 2005,~\citet{mihalcea2005making} introduced a dataset of humorous and non-humorous one-liner sentences sourced from the web alongside machine learning approaches to predict whether a sentence is humorous or not. Another dataset often used is the \emph{Pun of the Day} dataset created by~\citet{yang2015humor}. This data set consists of puns scraped from a website and non-humorous data acquired from, amongst others, news websites. Recently, social media data have become a popular source for textual humor datasets, with datasets based on tweets~\cite{potash2017semeval, castro2018crowd, khandelwal2018humor} and Reddit posts~\cite{weller2019humor} being proposed. While most available datasets are in English, corpora in Italian~\cite{reyes2009analysis}, Spanish~\cite{castro2018crowd}, Chinese~\cite{chen2018humor},  and Russian~\cite{ermilov2018stierlitz, blinov2019large} have been created. Typically, these datasets come with binary labeling, \ie, texts are annotated on whether they are humorous or not. This is different in, \eg, \emph{\#HashtagWars}~\cite{potash2017semeval} and the dataset by~\citet{castro2018crowd}, where texts are annotated with different degrees of humor. For task 7 of SemEval-2017~\cite{miller2017semeval}, English puns have been labeled for the location and interpretation of the word responsible for the pun. Task 7 of Semeval-2021~\cite{meaney2021semeval} introduced a dataset in which social media posts are not only labeled for humorous intentions but also regarding controversy and offensiveness.
Another approach to creating text-based humor recognition datasets is headline editing~\cite{west2019reverse, hossain2020stimulating}.~\citet{west2019reverse} let annotators edit a funny news headline into a serious one, while~\citet{hossain2020stimulating} reverse this process, encouraging annotators to modify a serious headline into a funny one.

Earlier textual humor recognition approaches relied on handcrafted features such as antonymy or alliterations and machine learning models such as \acp{SVM} and Random Forests~\cite{mihalcea2005making, yang2015humor}. In recent years, \acp{DNN} have been employed. To give an example,~\citet{chen2018humor} build a \ac{CNN}, while~\citet{ren2022attention} propose an approach based on \acp{LSTM} and an attention mechanism. With the advent of pretrained transformer-based \acp{LLM} like BERT~\cite{devlin2019bert}, the focus of textual humor detection shifted towards such models, motivated by their promising performance in many \ac{NLP} downstream tasks. For example,~\citet{weller2019humor} finetune BERT for the recognition of jokes. Similarly,~\cite{annamoradnejad2020colbert} make use of BERT sentence embeddings to tell jokes and non-jokes apart. 
\revision{
In~\cite{chen2024talk}, a framework and dataset towards training generative \acp{LLM} to respond in a humorous way is proposed.

More recent and more extensive \acp{LLM} such as LLaMa~\cite{Touvron2023LLaMAOA} or GPT-4~\cite{Achiam2023GPT4TR} have received considerable attention from the \ac{NLP} community, in particular due to their impressive few-shot and zero-shot capabilities. In this context,~\citet{jentzsch-kersting-2023-chatgpt} investigate to what extent ChatGPT is able to generate and explain jokes. Similarly,~\citet{baranov-etal-2023-told} test ChatGPT and Flan-UL2~\cite{Tay2022UL2UL} w.r.t. detection and continuation of jokes.

Another line of research related to textual humor recognition is concerned with memes, i.e., humorous combinations of images and text. To give but a few examples,~\citet{liu-etal-2022-figmemes} computationally analyze figurative language in memes, \citet{pramanick-etal-2021-momenta-multimodal} address the problem of detecting harmful memes, and ~\citet{pranesh2020memesem} propose a method to determine the sentiment of memes.}

\subsection{Multimodal Humor recognition}\label{ssec:rw_multi}
The fact that scholars have made promising attempts to model humor based on texts alone seems problematic as semantics is only one aspect of an utterance. Especially when it comes to more natural scenarios such as dialogues between people, voice and mimic often complement the semantic content of an utterance and may help both humans and machine learning models to grasp more validly whether an utterance is meant to be humorous or not. Interestingly, in this regard, the investigation of multimodal approaches to humor recognition gained traction considerably later than text-only humor recognition. To the best of our knowledge,~\citet{bertero2016deep} were the first to create a multimodal dataset for humor detection in 2016. They extracted dialogues from the TV sitcom \emph{The Big Bang Theory} and utilized the canned laughter to automatically label them as being humorous or not. Similar datasets, also based on \emph{The Big Bang Theory} and labeled using canned laughter, are created in~\cite{patro2021multimodal},~\cite{kayatani2021laughing}, and~\cite{alnajjar-etal-2022-laugh}. In general, TV shows are a popular source for multimodal humor datasets.
~\citet{wu2021mumor} introduce two datasets MUMOR-EN and MUMOR-ZH, constructed from the English sitcom \emph{Friends} and a Chinese sitcom, respectively. MaSaC~\cite{bedi2021multi} consists of Hindi-English code-mixed sitcom dialogues manually annotated for the presence of humor as well as sarcasm. A different approach is presented in~\cite{yang2019multimodal}, where the authors obtained humor labels by exploiting time-aligned user comments for videos on the Chinese video platform Bilibili.~\citet{hasan2019ur} compile their dataset UR-Funny from TED talk recordings, using laughter markup in the provided transcripts to automatically label punchline sentences in the recorded talks. For Open Mic,~\citet{mittal2021so} collected standup comedy recordings and used the audience's laughter to create annotations indicating the degree of humor on a scale from zero to four. Similar to text-only datasets, most multimodal datasets are in English, notable exceptions being the already mentioned MUMOR-ZH~\cite{wu2021mumor}, MaSaC~\cite{bedi2021multi}, the Chinese dataset used in~\cite{yang2019multimodal} and M2H2~\cite{chauhan2021m2h2}, which is based on a Hindi TV show.

Regarding multimodal humor prediction,~\citet{bertero2016deep} experiment with \acp{CNN} and \acp{RNN}, but also \acp{CRF}.\revision{~\citet{yang2023multi} utilize \acp{RNN} for processing sequences of body pose and facial expression features and combine them with BERT encodings of verbal utterances.}
Most recent approaches often use (self-)attention mechanisms to fuse representations of different modalities. For instance, the MuLOT model proposed by~\citet{pramanick2022multimodal} employs attention within and across different modalities.~\citet{hasan2021humor} obtain Transformer-based embeddings for the audio, video, and text modality and complement the latter with additional textual features inspired by humor theories. Further examples include the models introduced in~\cite{hasan2019ur},~\cite{bedi2021multi},~\cite{kayatani2021laughing}, and~\cite{patro2021multimodal}. \revision{\citet{chauhan-etal-2022-sentiment} also employ a Transformer architecture and, in addition, exploit sentiment and emotionality predictions as auxiliary tasks for humor detection.}. The aforementioned methods were developed specifically for the humor detection task. Besides, models intended for multimodal sentiment-related tasks in general have been introduced and evaluated also for the humor detection task recently. Examples of such generic approaches are \ac{MulT}~\cite{tsai2019multimodal}, \ac{MISA}~\cite{hazarika2020misa}, and \ac{BBFN}~\cite{han2021bi}, which have all shown promising performance for the UR-Funny dataset~\cite{hasan2019ur}.

A variant of \ac{Passau-SFCH} was already featured in the \textsc{MuSe-Humor} subchallenge of the \ac{MuSe} 2022~\cite{Christ22-TM2, Amiriparian22-TM2}. Here, participants were tasked with predicting the presence of humor in \ac{Passau-SFCH}. Participants could utilize the text, audio, and video modality. Several systems were proposed to tackle this task, all of them employing the Transformer architecture~\cite{xu22hybrid,chen22integrating, li22hybrid}. Key to this data set, as well as the one presented here is that it addresses focal shortcomings in the extant approaches mentioned above. First, the detection of humor occurs independent of laughter, which humor scholars have long described as theoretically orthogonal to humor, such that laughter-based recognition can lead to false positive and, especially, false-negative errors in humor recognition~\cite{attardo2019humor}. Second, the extant methodologies do not account for the notion long established in social-psychological research on humor that humor is multifaceted, i.e., involves substantially different types which can induce fundamentally different intra- and interpersonal affective, cognitive, and conative responses, e.g. ~\cite{avolio1999funny, huo2012only, cooper2018leader}.

\section{Passau SFCH Database}\label{sec:data}
The \ac{Passau-SFCH} database was originally created to develop a multimodal measurement of executive humor, \ie, humor utilized by corporate leaders. 
% Lukas: added this again
Football or professional sports in general, taking coaches as executives, provides a suitable and previously used context~\cite{day2012sporting} for this purpose.
%for at least two reasons.

From the methodological point of view, there are at least two reasons for choosing recordings of football press conferences for the purpose of automatic humor recognition: 
%First, the amount of publicly available data is unmatched. 
First, a large amount of relevant data, including audiovisual recordings of press conferences, is publicly available.
Second, football press conferences are quasi-experimental settings~\cite{wolfe2005sport} including ad-hoc communication. 
%which facilitates comparability across coaches and time.
Football coaches and journalists meet on a regular basis
%-- with pre-match conferences a couple of days prior to the match and post-match conferences just after the match -- 
and follow standardized, structured procedures and rules, such as requirements for positioning of lights and a predetermined protocol of opening statement and question \& answer format~\cite{DFL_Richtlinien}. The standardized setting facilitates comparability across individual videos, different coaches, and time.

\Crefrange{ssec:collection}{ssec:stats} introduce the \ac{Passau-SFCH} database in detail. Moreover, we discuss its differences from existing humor detection datasets in ~\Cref{ssec:comparison}
%and its limitations in~\Cref{ssec:limitations}.

\subsection{Collection}\label{ssec:collection}
With permission of the Deutsche Fußball Liga (DFL), we collected 59 pre-match press conference videos of 10 Bundesliga coaches from the first 13 match days of the season 2017/2018. All videos were publicly available on Facebook or YouTube. 
% Out for revision, because reviewer requested trimming <<<<
%and were downloaded from the different clubs' websites and social media presences. 
%We ensured that videos only comprise the actual press conference, manually removing advertisements or waiting time without coaches and the like, if necessary.
% >>>>
We aimed for our raw dataset to comprise at least 1 hour of communication for each of the 10 coaches. 
% out...
%The number of press conferences per coach ranges from 3 to 11
%because of difference in total press conference duration across clubs. 
Throughout this work, we refer to individual coaches by IDs from 1 to 10 for the sake of privacy.
As an important limitation, the coaches are all male, aged between 29--52\,years (average 42\,years), and of a similar cultural background. 
%We picked coaches from a range of clubs that represent the full spectrum of performance levels in the German Bundesliga, from top clubs competing in international tournaments to struggling teams, at danger of relegation to the second division. Similarly, we aspired to pick a sample with varying representations of professional experience, ranging from little to extensive experience as a coach, with the least experienced coach having coached less than 10 professional matches compared to more than 800 for the most experienced one. 

\Cref{tab:ds_stats} displays statistics on the lengths of the individual recordings and recordings per coach.
Importantly, our sample only includes native German speakers with full professional proficiency to mitigate the potential influence of different cultural contexts, a coach's preference for certain communication content and style as well as the audience's reaction to their humor~\cite{romero2006use}.

\begin{table}[h!]
    \centering
\resizebox{0.75\columnwidth}{!}{
    \begin{tabular}{lrrr}
    \toprule
    &\# & $\mu$ (Duration) & $\sigma$ (Duration) \\
    \cmidrule{2-4}
    subjects & 10 & 01:05:30 & 00:15:48 \\
    videos & 59 & 00:11:06 & 00:06:51  \\
    \bottomrule
    
    \end{tabular}

    }\caption{Key dataset statistics. Durations are given in the hh:mm:ss format. All information refer to the data after parts where the coach is not speaking have been removed (cf.~\Cref{ssec:preprocessing}).}\label{tab:ds_stats}\end{table}

\subsection{Transcriptions}\label{ssec:transcriptions}
We obtain transcriptions for every video in four steps. First, we generate automatic transcriptions via an XLSR-Wav2Vec2 model~\cite{conneau2019unsupervised} fine-tuned on the German part of CommonVoice~\cite{commonvoice:2020}\footnote{\href{https://huggingface.co/jonatasgrosman/wav2vec2-large-xlsr-53-german}{https://huggingface.co/jonatasgrosman/wav2vec2-large-xlsr-53-german}}. 
Especially depending on the degree of the dialect spoken by the subjects, we find the results to be of mixed quality.

Hence, in the second step, the automatic transcriptions were corrected manually. Third, in order to be able to align the transcripts to the videos, we use the Montreal Forced Aligner tool~\cite{mcauliffe2017montreal} to create timestamps on the word level. Lastly, since we need the transcripts to be segmented into sentences for textual feature extraction, we apply the Transformer-based multilingual punctuation restoration model introduced by~\citet{guhr-EtAl:2021:fullstop}\footnote{\href{https://huggingface.co/oliverguhr/fullstop-punctuation-multilang-large}{https://huggingface.co/oliverguhr/fullstop-punctuation-multilang-large}} to the manually corrected transcriptions. 

\subsection{Annotation}\label{ssec:annotation}
 %BS: had to add "years" twice here - please check and keep adding units :)
% moved to appendix
%\revision{appendix? $>>>$}The annotators received a thorough preparation through mandatory, full-day training, and an illustrative handbook. In the training, the annotators gained a theoretical understanding of humor and were provided with practical annotation examples.\revision{$<<<$}

\begin{figure}[h!]
    \centering
    \includegraphics[width=\columnwidth, page=1]{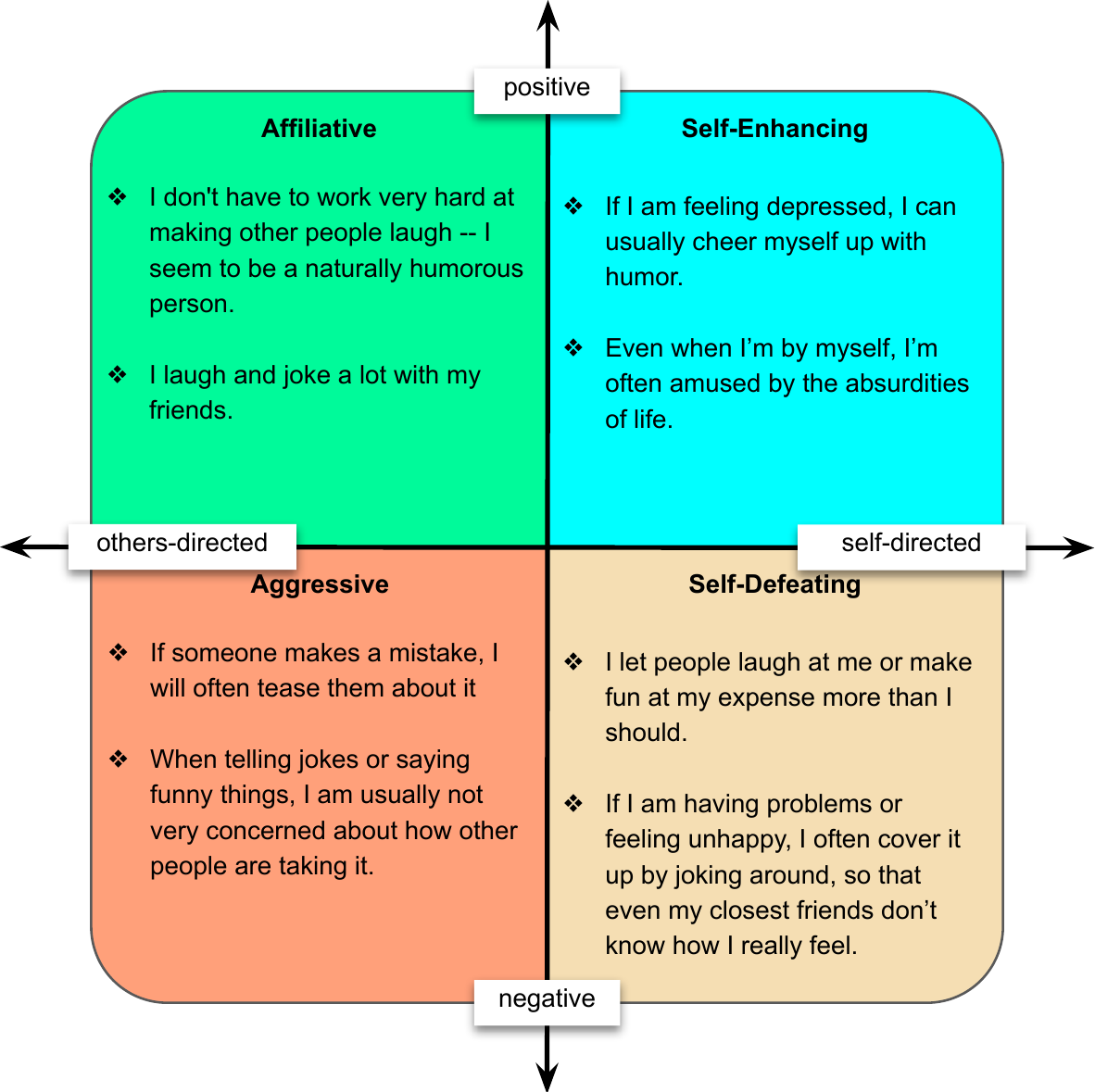}
    \caption{Illustration of the humor styles proposed by~\citet{martin2003individual}. The x-axis represents the direction dimension (self vs other-directed), and the y-axis the sentiment dimension (negative vs positive sentiment). The resulting quadrants each correspond to a humor style and are illustrated with items from the \ac{HSQ} which are associated with the respective humor style (cf.~\citet{martin2003individual} p.\,58f.)}\label{fig:humor_example}
\end{figure}

%BS: vs (not vs.) - I changed throughout
As annotation scheme, we choose the two dimensions of humor of the \ac{HSQ} proposed by~\citet{martin2003individual} -- arguably the most widely used taxonomy of humor in the social sciences, e.g. ~\cite{romero2006use, mesmer2012meta, robert2016impact, zeigler2020encyclopedia}. The HSQ distinguishes different styles of humor usage along their sentiment -- \ie, negative vs positive -- and their direction -- \ie, others-directed vs self-directed. This results in 4 different humor styles, as illustrated in~\Cref{fig:humor_example}. The \ac{HSQ} is particularly well suited for collecting real-time continuous annotations, given its comprehensiveness and conceptual parsimony and because it clearly refers to behavioral expressions of humor (cf.~\Cref{fig:humor_example}). Notably, in line with recent works in the social sciences, e.g. ~\cite{miron2022think}, we apply the HSQ typology to measure behavioral, individual, and real-time expressions of humor rather than trait humor. Examples of the different humor styles based on the gold standard (cf.~\Cref{ssec:gold_standard}) can be found in the supplementary material.

%BS: The figure seems to be a bitmap, even though it's a pdf, as it does not scale up. Please make sure it's drawn by us and scales up - thanks!

All videos are annotated by the same 9 trained annotators (5 female, 4 male) of similar age (average 25.33\,years; range 23-31\,years) and background. We utilize the DARMA~\cite{girard2018darma} software to obtain 2\,Hz annotation signals for both sentiment and direction via a joystick. Details on the annotation process are provided in the supplementary material.
%\revision{appendix? $<<<$}The annotation process was split into an initial screening of the press conference videos, and humor-style annotation using DARMA~\cite{girard2018darma}. With a rate of $2$\,Hz, the annotations are practically real-time. In line with the HSQ, one axis was dedicated to humor direction, the other to sentiment. The annotators were instructed not to cause any amplitude when no humor is present. Moreover, the annotators marked the segments in which the coach is speaking by pressing a button on the joystick which helps split the data into the relevant speaking time. Also, the annotators were able to pause the annotation process to ensure leeway for more difficult instances or any distractions.\revision{$>>>$} 

\subsection{Preprocessing}\label{ssec:preprocessing}

%utilizing the button annotations, 
We remove all parts of the video in which the coach is not speaking. This is necessary because we aim to predict humor solely based on data of the person in question, without considering possible reactions from the audience, especially laughter. After that, we are left with a dataset containing about 11 hours of video in total, with about an hour of video recordings per coach.

Regarding the annotation signals, we first apply thresholding to remove small, evidently erroneous amplitudes. Afterward, the signals are clipped to account for outliers and subsequently min-max normalised. An annotation value is considered an outlier if it deviates from the mean by more than $2.5$ standard deviations. We perform all these operations separately for each annotator and dimension (sentiment, direction). Moreover, because of the sparsity of segments considered humorous, we entirely ignore zero values in all of the computations mentioned above.

\subsection{Agreement}\label{ssec:agreement}

The agreement is reported for both the sentiment and direction annotations. Moreover, agreement on whether humor was detected at all can be computed. 

On average, about $4.4\,\%$ of the sentiment annotations per rater are non-zero, with a standard deviation of $2.3\,\%$. Similar values can be observed for the direction labels, with a mean of $4.7\,\%$ and a standard deviation of $2.1\,\%$. 
% OUT FOR REVISIION
%Because of this sparsity, we report the agreement only for the subset of the signals in which at least one annotator generated an amplitude for the respective dimension. For the agreement regarding sentiment and direction annotations, the \ac{CCC}
%TODO: equation? , as given in Equation~\ref{eq:ccc}, 
%is applied to the continuous signals. Comparing pair-wise different annotators, we obtain a mean pairwise \ac{CCC} of $.0985$ for sentiment and a mean pairwise \ac{CCC} of $.0845$ for direction. The standard deviations are $.0485$ and $\pm .0392$, respectively. 
We calculate the agreement for the sentiment and direction annotations by simplifying both of them to predictions of three classes, namely \textit{\{others-directed, self-directed, none\}} and \textit{\{negative, positive, none\}}. This way, we acquire for every 500\,ms time step and annotator a discrete prediction of sentiment and direction. On these predictions, Krippendorff's $\alpha$ values of $.1565$ and $.1893$ for direction and sentiment, respectively, are obtained. We consider our annotations valid as these inter-rater agreement scores show overlap, even though they are also not indicative of broad consensus. In fact, we believe the relatively low inter-rater agreement concerning the two HSQ axes can be related to the complex and contextualized nature of humor.
In order to estimate the agreement on whether humor is present, we treat every annotation as humorous if there is a non-zero value for either sentiment or direction, regardless of its sign. 
For this binary scheme, the overall Krippendorff's $\alpha$ value is $.1862$.
%Since this yields a binary labelling, the \ac{IoU} metric can be utilized to compare the annotators in a pair-wise manner. This results in a mean pair-wise IoU of $.1554$ with a standard deviation of $.0433$.

%~\Cref{fig:humor_agreement} shows the mean pairwise \ac{IoU} values for the detection of humor per annotator and coach. 
Overall, the agreement for all annotations is considerably low. 
%This is probably due to the inherent difficulty of this task. 
Humor is only displayed occasionally by coaches during a press conference. Moreover, oftentimes this humor is, on the one hand, rather subtle, and, on the other hand, domain-specific, \ie, a basic understanding of football may not be sufficient to understand the evoked allusions and jokes. Furthermore, small time lags in the 2\,Hz annotations distort the agreement values. Another problematic aspect is that the type of humor may vary for different coaches. This intuition is supported by a closer look at the agreements for each coach.
%\sa{if we need more space, this paragraph can be shortened.}

\begin{figure}[h!]
    \centering
    \includegraphics[width=\columnwidth]{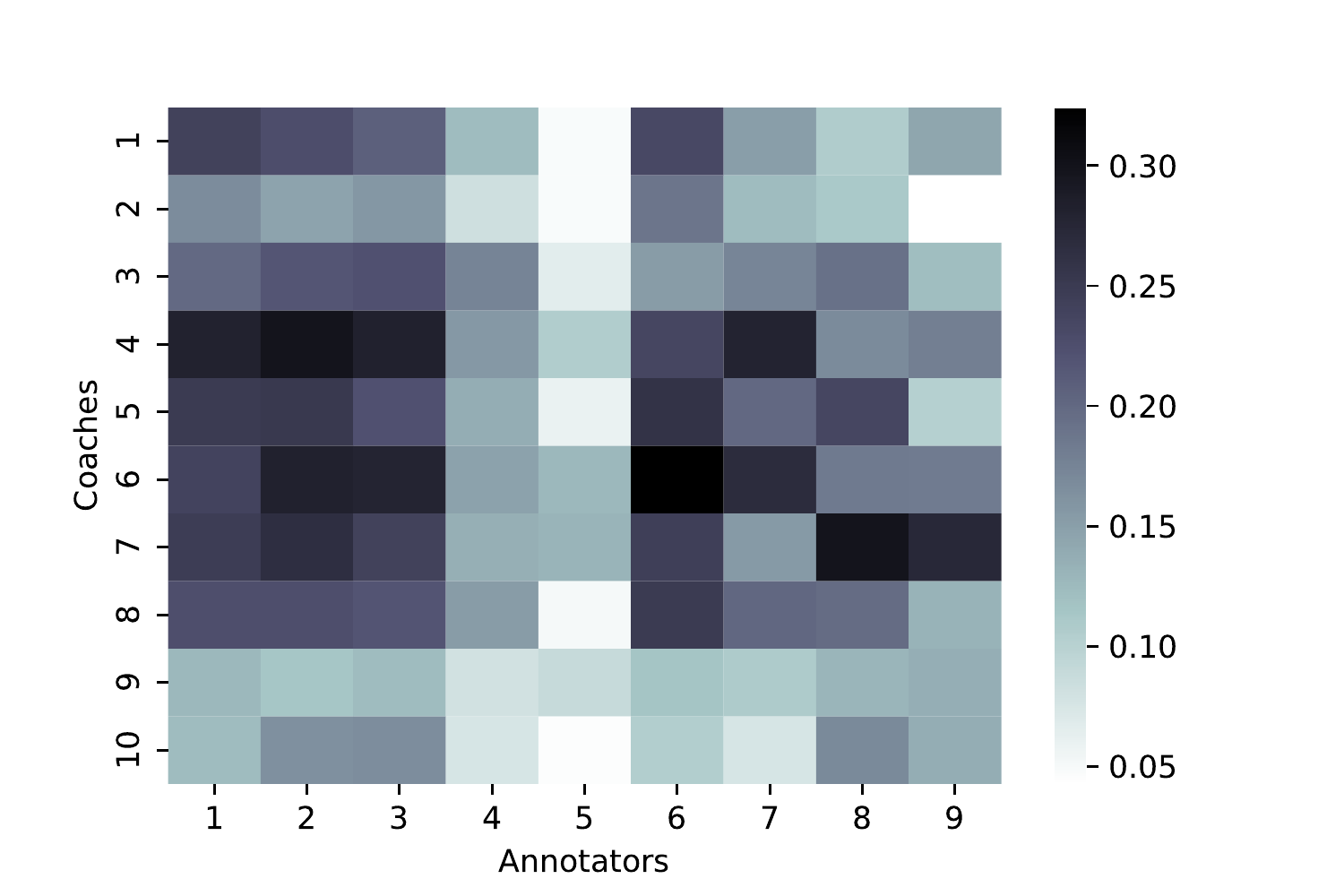}
    \caption{Mean humor agreements per coach and rater. Each cell represents the mean $\alpha$ value on the binary humor labels between an annotator and all the other annotators for the respective coach.}
    \label{fig:humor_agreement}
\end{figure}

As depicted in~\Cref{fig:humor_agreement}, the agreement is highly coach-dependent, with for example coach 9 displaying low mean $\alpha$ values compared to a relatively high agreement for, \eg, coach 6. Moreover, there are annotators who usually show a comparatively low agreement with other annotators, \eg, annotator 5. Depending on factors such as knowledge, norms, and other cognitive structures, people perceive different events as humorous~\cite{martin2006psychology}. In other words, our annotation data reflect that the expression and perception of humor are highly contextual~\cite{gloor2022risque}. With the gold standard creation process described in 
~\Cref{ssec:gold_standard}, we seek to reduce the influence of annotators that display low agreements. Our experiments show that \ac{ML} models can, to a degree, learn to distinguish between humorous and non-humorous instances of the dataset and between different types of humor (cf.~\Cref{sec:results}). This indicates that the proposed gold standard actually captures a distinct phenomenon, namely humor.

% Krippendorff's alpha is a metric for inter-annotator agreement in discrete classification settings. In order to make use of it, we map the signals for sentiment and direction to the three classes $\{-1, 0, 1\}$.   

% Table~\ref{tab:agreement} reports the mean pairwise agreements and the corresponding standard deviations for sentiment and direction.

% \begin{table}
%     \centering
%     \begin{tabular}{c|c|c|c}
%          Label & CCC & $\alpha$ & IoU \\ \hline
%          Sentiment & .0985 ($\pm.0485$) & .1061 ($\pm.0369$) & \\ 
%          Direction & .0845 ($\pm.0392$)& .0933 ($\pm.0370$& \\
%          Humor & - & & \\
%     \end{tabular}
%     \caption{Caption}
%     \label{tab:agreement}
% \end{table}

\subsection{Gold Standard Creation}\label{ssec:gold_standard} 
%\sa{this section should be shortened.}\lc{Shortened.}
We first generate gold standard labels for sentiment and direction of humor by fusing the respective annotations in every video. The following describes the gold standard creation procedure which was applied to both the sentiment and the direction annotations.

Let $A=\{a_1,...,a_9\}$ denote the set of annotators and $N=\{n_1,...,n_{|N|}\}$ the set of videos. We write $x_{n,a}$ for the signal corresponding to the annotation of a video $n\in N$ by a rater $a\in A$.  
Consistent with established practices for fusing continuous affect-related annotations~\cite{ringeval2017avec, stappen2021muse, stappen2021musetoolbox}, we utilize a fusion method in the fashion of \ac{EWE}~\cite{grimm2005evaluation}, \ie, we calculate agreement-based weighted sums from the individual annotators' signals. Since agreement varies depending on the coach 
% and dimension (sentiment, direction) 
in question (cf.~\Cref{ssec:agreement}), we calculate a weight $w_a(c)$ for every annotator $a$ and coach $c$. To do so, we only consider frames for which at least one annotator caused an amplitude. These signals are concatenated per annotator $a$ and coach $c$, resulting in signals $\mathbf{x}_{N\vert c,a}$.

Our weightings $w_a(c)$ then include two aspects of agreement. On the one hand, weights $w_a(c)'$ based on the agreements between the annotator $a$'s and all other annotators' signals are computed utilizing \ac{CCC}, as given in~\Cref{eq:w1}. On the other hand, we consider agreement regarding the sheer presence of humor. For example, if annotator $a$ disagrees with $a'$ on whether an utterance is self- or others-directed, they still agree on the utterance being humorous. To account for this kind of agreement, another weight $w_a(c)''$ is calculated based on the signals' absolute amplitudes, as can be seen in~\Cref{eq:w2}.

\begin{equation}\label{eq:w1}
    w_a(c)' = \frac{1}{|A|-1}\sum_{a'\neq a}CCC(\mathbf{x}_{N(c),a}, ~ \mathbf{x}_{N(c),a'}),
\end{equation}

\begin{equation}\label{eq:w2}
    w_a(c)'' = \frac{1}{|A|-1}\sum_{a'\neq a}CCC(|\mathbf{x}_{N(c),a}|, ~ |\mathbf{x}_{N(c),a'}|).
\end{equation}

In~\Cref{eq:w2}, $|\cdot|$ is meant to be an element-wise application of the absolute operator to a signal. The final weights $w_a(c)$ are then computed according to~\Cref{eq:weights}.

\begin{equation}\label{eq:weights}
    w_a(c) = \frac{w_a(c)' + w_a(c)''}{\sum_{a'\in A}(w_{a'}(c)' + w_{a'}(c)'')}.
\end{equation} 

Finally, the gold standard values for sentiment and direction are calculated as weighted sums.

Based on the fused annotations for sentiment and direction, a binary signal indicating the presence of humor can be computed. Any amplitude in any of the two dimensions is regarded as humor. 
To account for small lags, we further apply a windowing approach with frame size 2\,s and hop size 1\,s to the fused sentiment and direction signals, resulting in 39,682 segments. Motivated by low Jaccard agreements for binary humor detection (cf.~\Cref{ssec:agreement}), we drop the three annotators with the lowest mean inter-annotator agreement per video.  Each segment is labeled as humorous if at least 3 of the remaining 6 annotators have produced any amplitude for sentiment or direction within that segment. 

For consistency, we use the same 2\,s windowing approach to transforming the fused sentiment and direction signals. We reduce at most 4 values per window to one via mean pooling. Finally, all sentiment and direction segments are set to 0, if the corresponding humor label is 0.

\subsection{Dataset Statistics}\label{ssec:stats}
\begin{figure}[!h]
    \centering
    \includegraphics[width=.9\columnwidth]{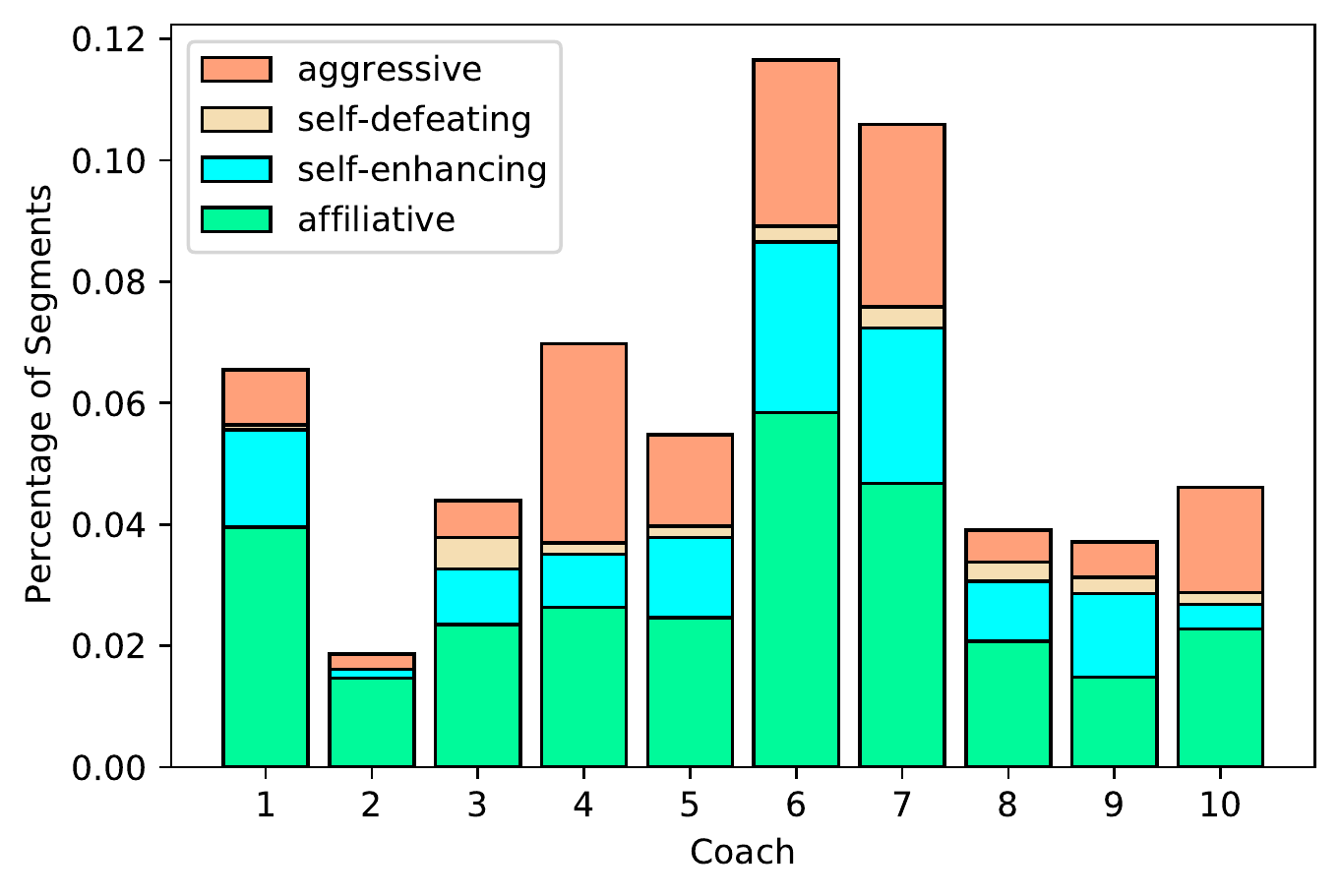}
    \caption{Percentage of humorous segments per coach in the gold standard, partitioned by humor style.}
    \label{fig:humor_bar}
\end{figure}

In total, $6.02$\,\% of the 39,682 segments are labeled as humorous. 
Different coaches display different amounts of humor in the recorded press conferences.~\Cref{fig:humor_bar} clearly illustrates considerable differences in the usage of humor in general. 

%To give an example, only about $2\,\%$ of coach 2's segments are labeled as humorous by the procedure described in~\Cref{ssec:gold_standard} while for coaches 6 and 7, more than $10\,\%$ of the respective utterances are considered humorous.\sa{The previous sentence can be removed.} 
%Furthermore,~\Cref{fig:humor_bar} breaks the humorous segments per coach down into the four different styles of humor (cf.~\Cref{fig:humor_example}).\sa{The previous sentence can also be removed.} 
For most coaches, affiliative -- \ie, positive, others-directed humor -- is the prevalent style of humor, with coach 4 being a notable exception. Self-defeating humor, on the contrary, is rare for every coach.
~\Cref{tab:distributions} provides further statistics about the distributions of different humor styles in the dataset. 

\begin{table}[]
    \centering
\resizebox{1\columnwidth}{!}{
    \begin{tabular}{llccc}
        \toprule
        & [\%] & \multicolumn{2}{c}{Sentiment} & \\
        & & negative & positive & $\Sigma$ \\ \midrule
         \multirow{4}{*}{Direction} & \multirow{2}{*}{self} & $.0427$& $.2229$ & $.2656$ \\ 
         & & ($.0443\pm .0338$) & ($.2084\pm .0845$) & ($.2528\pm .1022$) \\
         \cmidrule{3-5}
         
         & \multirow{2}{*}{other} & $.2560$ & $.4759$ & $.7327$ \\ & & ($.2337\pm .1113$) & ($.5109\pm .1123$) & ($.7457\pm .1031$) \\ \midrule 
         
         \multicolumn{2}{c}{\multirow{2}{*}{$\Sigma$}}  & $.2995$ & $.6996$ & \\ & &  ($.2784\pm .1070$) & ($.7205\pm .1068$) &  \\
         \bottomrule
    \end{tabular}
    }
    \caption{Humor style percentages, based on humorous segments only. Means and standard deviations refer to the different coaches.}
    \label{tab:distributions}
\end{table}

\subsection{Comparison with other Humor Datasets}\label{ssec:comparison}

\begin{table*}
\resizebox{2\columnwidth}{!}{
\begin{tabular}{llrrrlrl}
\toprule
     Authors (Dataset Name) & Extra Labels & \#Annotators & Dur.\,[h] & \#Spk. & Lang. & \%\,Humor & Scenario (scripted)\\
     \midrule
     
    ~\citet{bertero2016deep} ( - ) & - & - & ? & ? ($>$7) & en & 42.8 & \emph{Big Bang Theory} (yes) \\ 
     
    ~\citet{yang2019multimodal} ( - ) & - & - & 6.77 & 1 & zh & 26.72 & Bilibili platform (yes) \\
     
    ~\citet{hasan2019ur} (\textsc{UR-Funny}) & - & - & 90.23 & 1741 & en & 50 & TED Talks (yes) \\
     
    ~\citet{mittal2021so} (\textsc{Open Mic}) & - & 3 & {\textasciitilde } 17 & ? & en & 87.87 & Standup Comedy (yes) \\
     
    ~\citet{wu2021mumor}(\textsc{Mumor-ZH}) &- & 3 & 18.12 & 91 & zh & 28.36 & TV sitcoms (yes) \\
     
    ~\citet{wu2021mumor} (\textsc{Mumor-EN}) & -  & 3 & 9.03 & 259 & en & 24.59 & TV sitcoms (yes)\\
     
    ~\citet{bedi2021multi} (\textsc{MaSaC}) & Sarcasm & 3 & 80.98 & 5 & hi/en & 37.9 & TV show (yes) \\
     
    ~\citet{patro2021multimodal} ( - ) & - & - & 84 & ? ($>$7) & en & 82.5 & \emph{Big Bang Theory} (yes) \\
     
    ~\citet{kayatani2021laughing} ( - ) & - & - & 77.7 & 10 & en & $>$25 & \emph{Big Bang Theory} (yes) \\ 
     
    ~\citet{chauhan2021m2h2} (\textsc{M2H2}) & - & 3 & 4.5 & 41 & hi & 33.74 & TV series (yes) \\

    ~\citet{chauhan-etal-2022-sentiment} (\textsc{SHEMuD}) & Sentiment, Emotionality & 4 & 4.5 & 41 & hi, en & 33.74 & TV series (yes) \\
     
     \midrule
     \multirow{2}{*}{\textbf{Ours (\ac{Passau-SFCH})}} & Sentiment and & \multirow{2}{*}{9} & \multirow{2}{*}{10.92} & \multirow{2}{*}{10} & \multirow{2}{*}{de} & \multirow{2}{*}{6.02} & \multirow{2}{*}{Press Conferences (no)} \\
     & Direction~\cite{martin2003individual} & & & & & \\
     
     \bottomrule
\end{tabular}
}
\caption{Comparison of existing multimodal (audio, text, video) humor detection datasets and \ac{Passau-SFCH}. \emph{Dur.} denotes the overall duration of each dataset, \emph{\#Spk} the number of speakers in it. Regarding the language abbreviations, \emph{zh} stands for Chinese, \emph{hi} for Hindi and \emph{de} for German. Note that the \textsc{Open Mic} dataset by~\citet{mittal2021so} is special since it contains humor intensity ratings on a Likert scale (0-4) and uses recordings of TED talks as negative examples. \revision{\textsc{SHEMuD} is an extension of \textsc{M2H2} with additional translations and labels.}} \label{tab:comparison}

\end{table*}

In contrast to existing humor detection datasets, the subjects in our database act, in general, spontaneously, as they do not know in advance the exact questions they will be asked at the press conference. 
However, it can be argued that professional football coaches are used to handling such situations. Hence, it is reasonable to speak of a \emph{semi-staged} situation here, in contrast to fully staged scenarios like acted TV shows, standup comedy, or TED talks. 

Besides, our dataset is the first to include annotations according to the \ac{HSQ}. \revision{\textsc{SHEMuD}~\cite{chauhan-etal-2022-sentiment} also comes with a sentiment label, corresponding to one of the two \ac{HSQ} dimension.} 
While other datasets are labeled automatically, for example by using canned laughter~\cite{bertero2016deep, patro2021multimodal}, or by three human annotators~\cite{wu2021mumor, chauhan2021m2h2}, each video in our database has been explicitly labeled for humor by the same 9 annotators. In addition, to the best of our knowledge, \ac{Passau-SFCH} is the first dataset for both textual-only and multimodal humor detection in German.

Humor is distributed comparably sparsely across the recordings in \ac{Passau-SFCH}, with only $6.02\,\%$ of the segments being labeled humorous. In other datasets, at least $24\,\%$ of the annotated units are considered humorous. This sparsity is due to the rather matter-of-fact nature and objectives of press conferences. 

In~\Cref{tab:comparison}, we provide a detailed comparison of our dataset with other humor datasets. An aspect not covered by~\Cref{tab:comparison} is the annotation level. Different from all existing humor databases except the one created by~\citet{yang2019multimodal}, \ac{Passau-SFCH} is labeled in a time-continuous manner. All other datasets listed in~\Cref{tab:comparison} are annotated at utterance level, \ie, an utterance is either a punchline/joke or not. In \ac{Passau-SFCH}, also due to the comparably natural setting, an utterance may consist of humorous and non-humorous segments.

\section{Experimental Setup}\label{sec:experiments}

\begin{figure*}[h!]
    \centering
    \includegraphics[width=.95\linewidth, page=1]{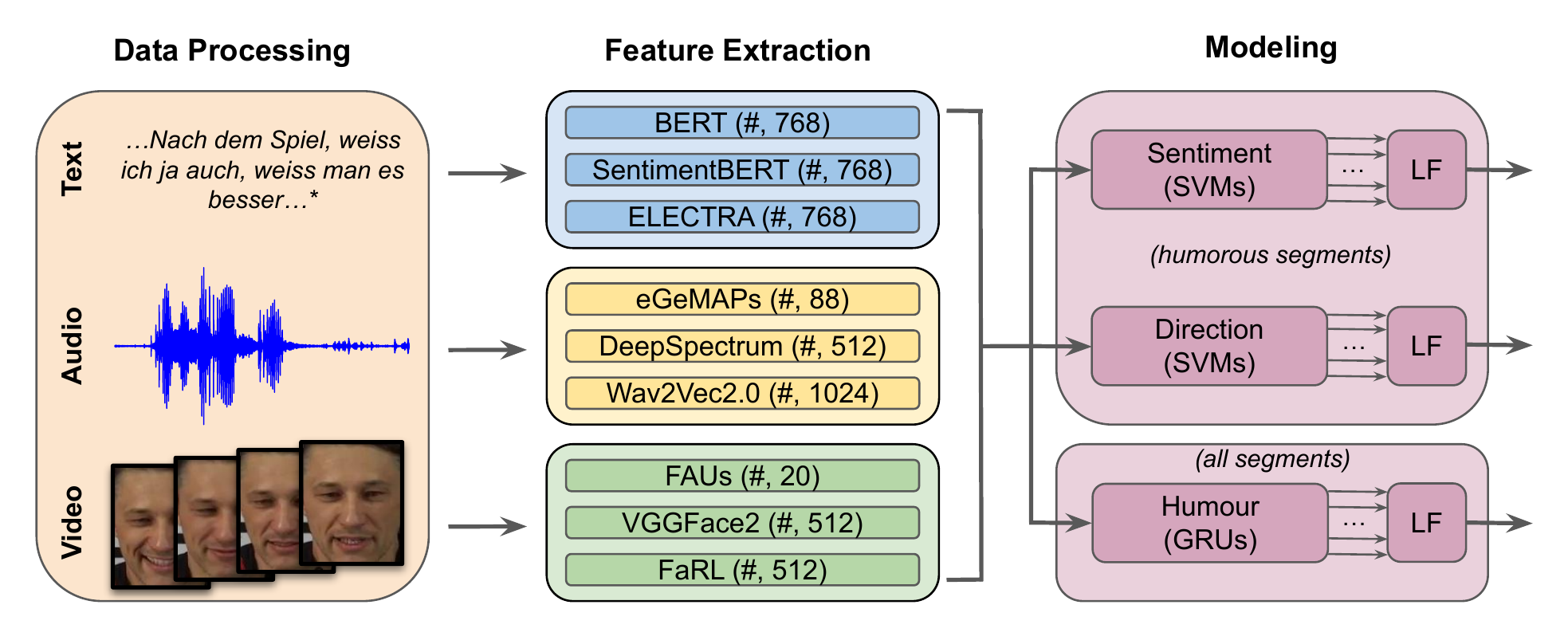}

    \caption{Overview of the setup of the unimodal and corresponding late fusion experiments. For all segments, three features for each of the three modalities are extracted and fed into unimodal systems for Sentiment Prediction, Direction Prediction,  and Humor Recognition, respectively. In, addition, late fusions (LF) are conducted. *The sentence roughly translates to \textit{``After the game, I know, one knows better''.}}\label{fig:system}
\end{figure*}

Our experiments are illustrated in~\Cref{fig:system}. In order to explore the strengths and weaknesses of all three available modalities, we compute three different contemporary feature sets for each of them. We then address the tasks of humor recognition as well as sentiment and dimension prediction. For the latter two, \acp{SVM} are utilized, while we employ \acp{GRU} for the recognition of humor. Given their simplicity and capacity to model sequences, \acp{GRU} are a natural choice for the task at hand. Note that for the experiments regarding sentiment and direction, only the segments labeled as humorous are considered. We opt for \acp{SVM} for these two tasks after we observed massive overfitting with \acp{GRU} on the resulting comparatively small datasets. 
% Out for revision
%The focus of our experiments lies on unimodal approaches, but we also conduct simple late fusions of trained models, gaining insight into the benefits of combining different modalities. 
In order to study the different modalities' and features' suitability for predicting humor, sentiment, and direction, we focus on simple, unimodal experiments. In addition, we experiment with combining the different modalities utilizing both a basic late fusion technique and more advanced multimodal training methods.

\subsection{Features}\label{ssec:features}
Given the multimodal nature of humor, we make use of audio, video, and text data.
For each of the three modalities, three different feature sets are computed, using both Transformers and more `conventional' machine learning techniques.
\subsubsection{Audio}
Regarding audio, we opt for the \ac{egemaps}~\cite{eyben_geneva_2016}, \ac{DS}~\cite{Amiriparian17-SSC},  and \ac{W2V}~\cite{baevski2020wav2vec}. All extracted audio features are aligned to the initial $2$\,Hz timestamps due to step sizes of $500$\,ms.

The \ac{egemaps} feature set comprises 88 speech-related paralinguistics features such as pitch, loudness, and \acp{MFCC}. It has been successfully applied in \ac{SER} before~\cite{wagner2018deep,amiriparian2019deep,baird2019can}, and is thus a promising approach to affect-related tasks such as humor detection. The \ac{egemaps} features are extracted for every 2\,s segment via  \opensmile{}~\cite{eyben2010opensmile}\footnote{\href{https://github.com/audeering/opensmile}{https://github.com/audeering/opensmile}} using the standard configuration with a frame size of 1\,s and a step size of 500\,ms. 

\ds\footnote{\href{https://github.com/DeepSpectrum/DeepSpectrum}{https://github.com/DeepSpectrum/DeepSpectrum}}~\cite{Amiriparian17-SSC}, on the contrary, is a \ac{DNN}-based toolkit designed to extract high-dimensional features from audio data. It has demonstrated its suitability for tasks related to ours such as \ac{SER}~\cite{Ottl20-GSE,amiriparian2019deep,amiriparian2022deepspectrumlite} and sentiment analysis~\cite{Amiriparian17-SAU}. To compute the features, (Mel-)spectrograms or chromagrams are fed into \acp{CNN} pretrained on image data. In our experiments, we opt for a 128-band Mel-spectrogram colored according to the \textit{viridis} colormap as a visual representation of the audio data. A \textsc{DenseNet121}~\cite{huang2017densely} pretrained on the ImageNet~\cite{russakovsky2015imagenet} dataset is chosen as the \acs{CNN} feature extractor. The feature vector is then given by the $1024$-dimensional output of its last pooling layer. We utilize a window size of $1$\,s and a frame size of $500$\,ms. 

Recently, Transformers~\cite{vaswani2017attention} have emerged as an alternative paradigm for generating high-dimensional audio features. Such models, \eg, \ac{W2V}~\cite{baevski2020wav2vec} or \textsc{HuBERT}~\cite{hsu2021hubert} are pretrained on large amounts of speech data utilizing self-supervised tasks. Features can then be devised from the internal representations of the model. \ac{W2V} has shown promising performance in \ac{SER} tasks~\cite{wagner2023dawn, pepino2021emotion}. For our purposes, we choose the same fine-tuned XLSR-Wav2Vec2 model\footnote{\href{https://huggingface.co/jonatasgrosman/wav2vec2-large-xlsr-53-german}{https://huggingface.co/jonatasgrosman/wav2vec2-large-xlsr-53-german}}~\cite{conneau2019unsupervised} as for the transcriptions. XLSR-Wav2Vec2 is based on \ac{W2V} but pretrained on more than 100 languages. In order to obtain embeddings, we use sliding windows with a frame size of 2\,s and hop size of 500\,ms to feed audio into the pretrained model. We then average over the representations returned by the final layer.

\subsubsection{Text}\label{sssec:features_text}

For the textual features, we apply three pretrained Transformer models, BERT~\cite{devlin2019bert}, German SentimentBERT~\cite{guhr2020training}, and \textsc{ELECTRA}~\cite{clark2020electra}. 
\textsc{BERT}~\cite{devlin2019bert} has already been used in the context of both text-only~\cite{annamoradnejad2020colbert, weller2019humor} and multimodal humor recognition~\cite{han2021bi,patro2021multimodal}. Since the subjects of \ac{Passau-SFCH} speak German, we employ a German version of \textsc{BERT}\footnote{\href{https://huggingface.co/bert-base-german-cased}{https://huggingface.co/bert-base-german-cased}}. 
German SentimentBERT~\cite{guhr2020training}\footnote{\href{https://huggingface.co/oliverguhr/german-sentiment-bert}{https://huggingface.co/oliverguhr/german-sentiment-bert}} is a Transformer model trained on German sentiment analysis data. Since sentiment is one of the two dimensions of the humor proposed in the \ac{HSQ}, we hypothesize that German SentimentBERT might outperform the more generic models. 
\textsc{ELECTRA}~\cite{clark2020electra} is the third Transformer model which often outperforms the original \textsc{BERT} model in various \ac{NLP} tasks~\cite{clark2020electra}. For our experiments, we use a version of \textsc{ELECTRA} which is pretrained on German texts\footnote{\href{https://huggingface.co/german-nlp-group/electra-base-german-uncased}{https://huggingface.co/german-nlp-group/electra-base-german-uncased}}.

We carry out the same feature extraction procedure for all three models. An utterance is already segmented into sentences due to punctuation restoration. Sentence-level features are then obtained by averaging the last 4 layers' representations for the special \textsc{[CLS]} token, following~\cite{sun2020multi}. These features are aligned to the initial $500$\,ms annotations via the timestamps obtained from the \ac{ASR} output. The text feature for a $500$\,ms frame is the average overall -- in practice, at most 2 -- sentences intersecting with this frame. Besides, we also experimented with token-level features but found the results to be slightly worse than those for sentence-level embeddings.

\subsubsection{Video}
Because of the rather static press conference setting, we focus on face-related features for the video modality. This is further motivated by the coaches' faces typically being zoomed in on when the coaches are speaking. In the first step, we extract all faces using \textsc{MTCNN}~\cite{zhang2016mtcnn} with a hop size of $500$\,ms in order to be in line with the $2$\,Hz annotations. We then make sure to only include the faces of the actual coaches. This is achieved by an automatic step, comparing FaceNet\footnote{\href{https://github.com/timesler/facenet-pytorch}{https://github.com/timesler/facenet-pytorch}}~\cite{schroff2015facenet} embeddings of detected faces with embeddings of reference pictures. Subsequently, manual correction is carried out.
Based on the extracted faces, three different feature sets are computed.
\acp{FAU} describe facial expressions. Ekman identified 44 such units and also proposed mappings from their activation to emotional states~\cite{ekman1992facial}. As \acp{FAU} play an important role in manual and automatic affect detection~\cite{martinez2017automatic}, we extract estimations of 20 different \acp{FAU}' activations using the \textsc{Py-Feat} toolkit\footnote{\href{https://py-feat.org}{https://py-feat.org}}. 

Second, we extract latent facial features from a ResNet50~\cite{he2016deep} trained on the \vggf{}\footnote{\href{https://github.com/WeidiXie/Keras-VGGFace2-ResNet50}{https://github.com/WeidiXie/Keras-VGGFace2-ResNet50}} face recognition dataset. A $512$-dimensional feature vector is obtained by taking the representations of its last hidden layer. 

Third, we apply \ac{farl}~\cite{zheng2022general} which is a pretrained Transformer model designed for tasks related to facial features. Using the \ac{farl} image encoder we obtain $512$-dimensional features for the extracted face frames.
%\sae{\sout{In computer vision, pretrained Transformer models are becoming increasingly common, too. \ac{farl}~\cite{zheng2022general} is a Transformer model intended for tasks related to facial features. It is pretrained on a visual-linguistic and a low-level masked image modelling task utilizing a subset of the LAION~\cite{schuhmann2021laion} dataset. We construct $1024$-dimensional features for the extracted face frames via the pretrained \ac{farl} image encoder.}}

%\sae{\sout{All visual features are aligned to the initial 2\,Hz annotations since the faces are extracted at the same rate.}}

\subsection{Models}\label{ssec:unimodal_models}

Using the obtained feature sets from each modality, we train two sets of unimodal machine learning models, namely a \ac{GRU}-\ac{RNN} approach for the detection of humor segments and a \ac{SVM} approach for the recognition of sentiment and direction. For our multimodal approaches, we employ late fusion for all three prediction targets. For humor detection, we additionally explore a set of advanced methods for multimodal sentiment analysis, namely \ac{LMF}~\cite{liu-etal-2018-efficient-low}, \ac{MulT}~\cite{tsai2019multimodal}, and \ac{MISA}~\cite{hazarika2020misa}.

\subsubsection{GRU-RNNs for Humor Detection}\label{ssec:grus}
Since all features are extracted at a 2\,Hz rate, each 2\,s segment corresponds to up to 4 consecutive feature vectors.
\acp{RNN} such as \acp{GRU}~\cite{cho2014properties} are capable of modeling dependencies in sequential data and are thus a natural choice for this task.
For each of the 9 feature sets, we train 10 different \acp{GRU}, each time setting aside one coach as a test subject. This way, we can gain insights into the predictability of each coach's individual humor style.

Hyperparameter searches, informed by previous results in~\cite{Christ22-TM2, Amiriparian22-TM2}, regarding the number of layers, direction of the information flow (uni- or bidirectional), internal representation sizes, and learning rate of the training process are conducted for each feature set individually. In the training process, we split the data of the 9 training coaches randomly into a train and development set. Every model is trained for at most 20 epochs but potentially stopped early after 5 epochs of no improvement on the development set. To mitigate the influence of random initialization values, we repeat every training process 5 times with different, fixed random seeds. As a loss function, binary cross entropy is used. We choose AdamW~\cite{loshchilov2018decoupled} as the optimization algorithm. The \ac{AUC} is utilized as the evaluation metric and as the early stopping criterion. 

\subsubsection{Multimodal Training for Humor Detection}\label{ssec:train_multi}

In recent years, a plethora of methods dedicated to multimodal sentiment analysis based on audio, text, and video have been proposed. Since \ac{Passau-SFCH} comprises audio, video, and text data, we evaluate a few such methods for their suitability in detecting humor in the \ac{Passau-SFCH} recordings. 
We opt for \ac{LMF}~\cite{liu-etal-2018-efficient-low}, \ac{MulT}~\cite{tsai2019multimodal}, and \ac{MISA}~\cite{hazarika2020misa} and make use of their implementations provided by~\citet{yu2021learning}\footnote{\hyperlink{https://github.com/thuiar/MMSA}{https://github.com/thuiar/MMSA}}.

\begin{figure}[h!]
    \centering
    \includegraphics[width=.85\linewidth]{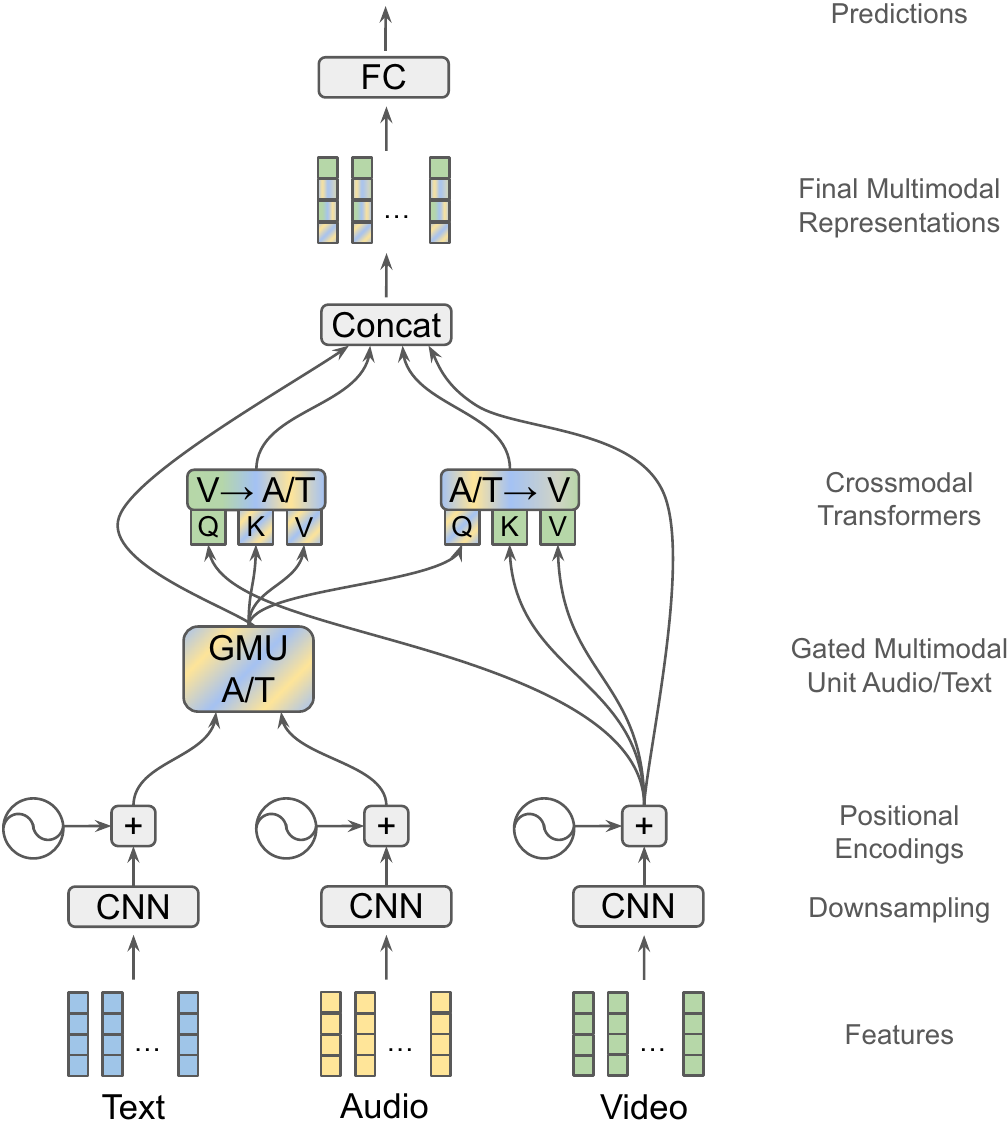}

    \caption{The proposed \ac{VFMM} architecture.}\label{fig:vfmm}
\end{figure}

\revision{Similar to the \ac{LSTM} approach, these three models only ingest representations of $2$\,s segments and predict the corresponding label. However, taking the entire clip instead of just $2$\,s into account can be expected to increase a model's performance, as a larger context of an utterance may include more cues for deciding whether an utterance is to be considered humorous or not. Hence, we also experiment with casting the humor problem as a sequence-to-sequence task. The Transformer-based \ac{MulT} model can be conveniently modified to predict the sequence of humor labels for a sequence of input features. We refer to this variant as \textit{\ac{MulT} -- full clips} in the following. 

We further propose \ac{VFMM}, a model architecture optimized for the problem at hand. Same as the full-clip \ac{MulT} model, it also processes entire clips. From the unimodal results (cf.~\Cref{tab:humor_results}), it is clear that the video modality is most promising for humor detection in our dataset. The experiments with \ac{MulT} show that the results do not tend to increase with more layers in the model. Informed by these findings, we design \ac{VFMM} as a compact architecture that gives priority to video data by design.~\Cref{fig:vfmm} illustrates the proposed \ac{VFMM} approach.}

\revision{More formally, the model takes in input feature sequences $t=t_1,...,t_T$, $a=a_1,...,a_t$, and $v=v_1,...,v_T$, corresponding to text, audio and visual representations. Same as in \ac{MulT}, \ac{CNN} layers $CNN_t$, $CNN_a$ and $CNN_v$ project them to a common size $d$, where $d$ is a hyperparameter.
\begin{equation}
    t' = CNN_t(t)
\end{equation}
\begin{equation}
    a' = CNN_a(a)
\end{equation}
\begin{equation}
    v' = CNN_v(v)
\end{equation}

Subsequently, fixed sinusoidal positional embeddings $PE = PE_1,...,PE_T$ are added to the representations, following the original Transformer approach~\cite{vaswani2017attention}.

\begin{equation}
    t'' = t' + PE 
\end{equation}

\begin{equation}
    a'' = a' + PE 
\end{equation}

\begin{equation}
    v'' = v' + PE 
\end{equation}

As audio and text are less promising than the visual modality, we fuse them into a compact combined $d$-dimensional representation $at$. This fusion is implemented via a \ac{GMU} module~\cite{arevalo2020gated}. 

\begin{equation}
    at = GMU(a'', t'')
\end{equation}
The representations $at$ are then combined with the visual feature sequence $v''$ by cross-modal encoder-only Transformers as employed in \ac{MulT}. The cross-modal Transformer variant considered here (denoted $CMT$) expects two sequences $s_1$ and $s_2$ of equal length. The sequence $s_1$ then serves as the query in the attention part of the Transformer, whereas $s_2$ accounts for the key and value sequences. Note that we only use one encoder layer per cross-modal Transformer here. There are two $CMT$ modules, as illustrated in~\Cref{fig:vfmm} and Equations \ref{eq:cross-modal1} and \ref{eq:cross-modal2}

\begin{equation}\label{eq:cross-modal1}
    v\rightarrow at = CMT(v'', at)
\end{equation}

\begin{equation}\label{eq:cross-modal2}
    at\rightarrow v = CMT(at, v'')
\end{equation}

The final multimodal representations are obtained by concatenating $v''$, $at$, $v\rightarrow at$, and $at\rightarrow v$, leading to a sequence of length $T$ and dimensionality $4\times d$. This sequence is fed to a fully connected layer with Sigmoid activation that outputs the predictions.

}

In all of the multimodal models, we opt for the best-performing features in the \ac{RNN} experiments as features for the three different modalities.
For all five approaches (\ac{LMF}, \ac{MISA}, \ac{MulT}, \ac{MulT} -- full clips, \ac{VFMM}), we conduct hyperparameter searches utilizing a speaker-dependent split of the dataset. We then train the models analogously to the \ac{RNN}-based experiments (cf.~\Cref{ssec:grus}) in a leave-one-subject-out manner. \revision{In the full-clip models, we observe benefits when restricting the attention matrices of the cross-modal Transformer modules. More precisely, we let each input element of a feature sequence attend only to at most $8$ neighboring elements to both the left and the right.}

\subsubsection{SVMs for Humor Style Classification}\label{sssec:svm_models}
Moreover, we investigate whether the sentiment and direction of humorous utterances can be recognized automatically. For these experiments, we only take humorous 2\,s segments into account. Since this limitation leaves us with only 2,387 data points, \acp{DNN} may not be a suitable choice. Instead, we train \acp{SVM} for both sentiment and direction prediction. We reduce both the sentiment and direction values to two classes $\{-1,1\}$, thus framing both tasks as classification problems. 

Because the features are extracted every 500\,ms, they must be reduced to the 2\,s label windows. This is achieved by computing the means of all feature values corresponding to a label window.

Analogously to~\Cref{ssec:grus}, hyperparameter search is performed for every feature set separately in a leave-one-subject-out manner, always setting aside one coach for validation. 
%BS: This is actually leave-one-subject-out and hence reproducible, so please CALL it also like that and not simply 10-fold CV :)
Subsequently, in line with~\Cref{ssec:grus}, 10 different models are trained for each feature set, each time leaving one coach out of the training data (leave-one-subject-out). We repeat every experiment with $5$ fixed seeds.

As the evaluation metric, \ac{AUC} is employed here as well. In order to obtain continuous predictions from \acp{SVM}, Platt Scaling~\cite{platt1999probabilistic} is applied. 

\subsubsection{Late Fusion}\label{sssec:late_fusion}
In order to explore the complementarity of different modalities, we conduct late fusion experiments for the recognition of humor as well as the prediction of both sentiment and direction. More specifically, we only consider the predictions obtained with the best-performing feature of each of the three modalities (text, audio, video) 
%BS: you kept shuffling the three - I try to always have text, audio, video :)
and build new predictions for every coach as follows. First, we z-standardize each feature's predictions. This is especially necessary for the predictions obtained by \acp{SVM} optimized for \ac{AUC}, as different \acp{SVM}' predictions may have different scales. Then, we compute a weighted sum where the weights are based on the performance during training. As for the \ac{SVM} experiments, we take the \ac{AUC} on the training set, reduced by $.5000$, \ie, chance level, as the weight. Regarding \acp{GRU}, the highest \ac{AUC} encountered on the validation set during the training process is chosen as the weight, also reduced by $.5000$. We explore all possible combinations of modalities, namely audio+text, audio+video, text+video, and audio+text+video.

\section{Results} \label{sec:results}
We report the results for both the humor detection experiments with \acp{GRU} in~\Cref{ssec:humor_results}. The \ac{SVM}-based experiments regarding the prediction of humor style and sentiment are discussed in~\Cref{ssec:style_results}. Furthermore, we compare the results of all experiments across the different coaches in~\Cref{ssec:style_comparison}.

\subsection{Humor Recognition}\label{ssec:humor_results}
\Cref{tab:humor_results} lists the results for the binary humor recognition task addressed with \acp{GRU}. All features lead to above-chance performance for every coach. However, the \ac{AUC} values differ considerably when comparing the three modalities. 

\begin{table}[h!]

\resizebox{1\columnwidth}{!}{
\centering
\begin{tabular}{lrrrr}
\toprule
Feature & \multicolumn{1}{c}{Mean (Std)} & \multicolumn{1}{c}{Min} & \multicolumn{1}{c}{Max} & \multicolumn{1}{c}{$>$ Chance} \\ \midrule \midrule

\multicolumn{5}{l}{\textbf{Audio}} \\
\ac{egemaps} & .7003 ($\pm$.0674) & .6037 & .8314
 & 100\,\%\\
 \ds & .6809 ($\pm$.0782) & .5935 & .8494 & 100\,\%\\
  \ac{W2V} & \textbf{.7505} ($\pm$.0687) & .6173 & .8741 & 100\,\%\\
  \midrule
  
  \multicolumn{5}{l}{\textbf{Text}} \\  
\textsc{BERT} & .7672 ($\pm$.0531) & .7126 & .8919 & 100\,\%\\
 \textsc{GerSent-BERT} & .7445 ($\pm$.0602) & .6525 & .8854 & 100\,\%\\
  \textsc{ELECTRA} & \textbf{.8044} ($\pm$.0689) & .7003 & .9169 & 100\,\%\\
  \midrule
  
\multicolumn{5}{l}{\textbf{Video}} \\
\acp{FAU} & .8361 ($\pm$.1022) & .5902 & .9509 & 100\,\%\\
 \vggf{} & .8349 ($\pm$.1048) & .6505 & .9480 & 100\,\%\\
  \ac{farl} & \textbf{\underline{.8962}} ($\pm$.0766) & .7038 & .9625 & 100\,\%\\

 \midrule

\multicolumn{5}{l}{\textbf{Late Fusion}} \\
A + T & .8283 ($\pm$.0679) & .7361 & .9321 & 100\,\%\\
 A + V & .8902 ($\pm$.0696) & .7424 & .9822
 & 100\,\%\\
  T + V & .9027 ($\pm$.0630) & .7734 & .9887 & 100\,\%\\
  A + T + V & .9038 ($\pm$.0600) & .7825 & .9903 & 100\,\%\\
   \midrule

\multicolumn{5}{l}{\textbf{Multimodal Training}} \\
\ac{LMF}~\cite{liu-etal-2018-efficient-low} & .8777 ($\pm$.0673) & .7432 & .9654 & 100\,\%\\
\ac{MulT}~\cite{tsai2019multimodal} & .8649 ($\pm$.0737) & .7521 & .9538 & 100\,\%\\
 \ac{MISA}~\cite{hazarika2020misa} & .8611 ($\pm$.0827) & .7140 & .9601
 & 100\,\%\\
 \ac{MulT} -- full clips~\cite{tsai2019multimodal} & .9180 ($\pm$.0595) & .7719 & .9850 & 100\,\%\\
 \ac{VFMM} -- full clips (ours) & \textbf{\underline{.9224}} ($\pm$.0555) & .7804 & .9847 & 100\,\%\\

\bottomrule
\end{tabular}
}\caption{Results for binary humor prediction with different features with \ac{AUC} as the evaluation metric. \emph{Mean} and \emph{Std} refer to the different coaches. More precisely, mean and standard deviations over the mean \acs{AUC} (5 seeds) per coach are reported (standard deviations for every single coach's 5 seeds are negligibly low in most cases). Analogously, \emph{Min} and \emph{Max} refer to the mean \acp{AUC} per coach. The column \emph{$>$ Chance} gives the percentage of coaches for which the mean \ac{AUC} was above chance level, \ie, .5. The best results per modality are boldfaced, and the best unimodal and multimodal results overall are underlined. }\label{tab:humor_results}
\end{table}

All \textbf{visual features}' means are higher than the mean value of any audio or text feature. One reason for the good performance of face-related features here may be that the coaches frequently smile or laugh during humorous utterances. It should be noted that laughter, with its meaning depending largely on the context of an utterance~\cite{curran2018social}, does not necessarily imply humor. In the press conference setting, however, it might be more correlated to humorous utterances than in-the-wild. Among the video-based approaches, \ac{farl} proves to be the best feature with a mean performance of $.8962$ \ac{AUC} and the lowest standard deviation across all coaches. The expert-designed \ac{FAU} features are outperformed by both deep learning-based approaches \vggf{} and \ac{farl}. 

\textbf{Textual features} fall behind the video-based ones with the best textual approach \textsc{ELECTRA} leading to a mean \ac{AUC} of $.8044$. This can partly be explained by the domain-specific nature of the texts in question. The textual features are extracted using Transformer models pretrained on more generic texts and thus may fail to capture football-specific humor. Another limiting factor for predicting humor in this setting using only the transcripts is that the transcripts are incomplete, as they only contain the coach's answers, but not the journalists' questions. The questions are, however, a crucial aspect of each utterance's context. Nevertheless, the performance of text-based features is, in general, more robust than for the other two modalities, with standard deviations across the coaches of at most $.0689$, namely for \textsc{ELECTRA}). Also, the lowest mean \acp{AUC} for an individual coach observed during the textual experiments, \ie, $.6525$ with \textsc{GerSent-Bert}, is higher than the lowest individual mean \ac{AUC} of $.5902$ encountered with visual features. 

The \textbf{audio modality} is less suitable for humor recognition, but still, all audio experiments result in \acp{AUC} over $.5000$. The performance of \ac{W2V} is even comparable with that of \textsc{Bert} and \textsc{GerSent-Bert}. However, it should be noted that due to its speech recognition pretraining task, \ac{W2V} embeddings also encode some linguistic information~\cite{triantafyllopoulos2022probing}. 
Other than in the video experiments, the handcrafted \ac{egemaps} feature set leads to a slightly better result than the non-Transformer deep learning method \dsns. 
Of note, for each modality, the most recent feature extraction technology -- all of them being based on Transformers -- performs best.   

\Cref{tab:humor_results} also demonstrates individual differences among coaches. It is clear from the reported standard deviations as well as minimum and maximum values that for some coaches, humor is harder to predict than for others. Notable discrepancies between the best and worst \ac{AUC} per coach can be observed for every feature set. The differences range from about $.1800$ \ac{AUC} for \textsc{BERT} to about $.3600$ for \acp{FAU}. Individual differences in predictability will be analyzed more thoroughly in~\Cref{ssec:style_comparison}.

The results for different late fusions prove that, to a degree, the three modalities are complementary. Fusing the best audio and text predictions outperforms the best text-only result with $.8283$ mean \ac{AUC} compared to $.8044$. The video modality can benefit from being combined with text only as well as audio and text simultaneously, the latter leading to a mean \ac{AUC} of $.9038$. Especially notable is the improvement regarding the minimum \ac{AUC} values per experiment. In the best unimodal case \ac{farl}, an \ac{AUC} of $.7038$ is achieved for the worst-performing coach in this experiment. For the worst performing coach in the three-modal fusion experiment, however, an \ac{AUC} of $.7825$ is observed, which marks a relative increase of more than $11\,\%$. 

The three advanced multimodal sentiment analysis methods \revision{based on $2$\,s segments (\ac{LMF}, \ac{MulT}, \ac{MISA})} do not outperform the results for late fusion of audio, text, and video. With mean  \ac{AUC} values of $.8777$, $.8649$, and $.8611$ for \ac{LMF}, \ac{MulT}, and \ac{MISA}, respectively, they even perform slightly worse than the unimodal \ac{GRU} for \ac{farl} features ($.8962$ \ac{AUC}). One reason for these comparably weak results may be that they only utilize the 2\,s segments, while all of these methods are designed with longer sequences such as in CMU-MOSI~\cite{zadeh2016mosi} or UR-Funny~\cite{hasan2019ur} in mind. Moreover, in our case, the video modality clearly trumps audio and text. Our weighted late fusion approach directly accounts for this imbalance, while the multimodal models may face difficulties to learn such information from the very short input sequences.

\revision{The models considering the full clips as inputs, however, prove to be superior over both the unimodal \acp{LSTM} and the late fusions. \ac{MulT} on the full clips accounts for a mean \ac{AUC} value of $.9180$, whereas our \ac{VFMM} approach is responsible for the best results overall, namely $.9224$ mean \ac{AUC}. Both models also exhibit lower standard deviations across subjects than all other multimodal methods, including late fusion. A comparison between the results of the $2$\,s segment \ac{MulT} version ($.8649$ mean \ac{AUC}) and the full clip \ac{MulT} ($.9180$ mean \ac{AUC}) suggests that the superior performance of the full clip models can largely be attributed to their consideration of larger contexts.

The results of our \ac{VFMM} model in comparison to the full-clip \ac{MulT} indicate that systematically giving preference to the visual modality is beneficial for the data at hand. In \ac{MulT}, each of the $6$ modality combinations ($V\rightarrow A$, $A\rightarrow V$, $T\rightarrow A$, $A\rightarrow T$, $V\rightarrow T$, $T\rightarrow V$) is processed by a corresponding cross-modal Transformer. Subsequently, the thus obtained representations are concatenated as the basis for prediction. This approach potentially introduces noise while losing valuable information from the video-only representations. \ac{VFMM}, in contrast, comes with only three modality fusions and retains the video-only representations via a residual connection. The overall number of trainable parameters of \ac{VFMM} is 270,785, which is less than $75$\,\% of the number of trainable parameters for the best \ac{MulT} configuration (369,409). 
}

\subsection{Humor Style Classification}\label{ssec:style_results}
We analyze the results of the experiments described in~\Cref{sssec:svm_models} for the discrimination of humor direction and sentiment in~\Cref{sssec:dir_results} and~\Cref{sssec:sen_results}, respectively.

\subsubsection{Direction}\label{sssec:dir_results}

\begin{table}[h!]
\resizebox{1\columnwidth}{!}{
\centering
\begin{tabular}{lrrrr}
\toprule
Feature & \multicolumn{1}{c}{Mean (Std)} & \multicolumn{1}{c}{Min} & \multicolumn{1}{c}{Max} & \multicolumn{1}{c}{$>$ Chance} \\ \midrule \midrule

\multicolumn{5}{l}{\textbf{Audio}} \\
\ac{egemaps} & \textbf{.5496} ($\pm$.1073) & .3964 & .7604 & 60\,\%\\
 \ds & .5396 ($\pm$.1178) & .2864 & .7656 & 80\,\%\\
  \ac{W2V} & .5314 ($\pm$.0896) & .4167 & .7135 & 60\,\%\\
  \midrule
  
  \multicolumn{5}{l}{\textbf{Text}} \\  
\textsc{BERT} & .6018 ($\pm$.1054) & .4363 & .7784& 80\,\%\\
 \textsc{GerSent-BERT} & .5756 ($\pm$.1449) & .2240 & .7424 & 80\,\%\\
  \textsc{ELECTRA} & \textbf{\underline{.6108}} ($\pm$.1081) & .4344 & .8281 & 90\,\%\\
  \midrule
  
\multicolumn{5}{l}{\textbf{Video}} \\
\acp{FAU} & \textbf{.5119} ($\pm$.0287) & .4697 & .5696 & 60\,\%\\
 \vggf{} & .5012 ($\pm$.1584) & .3489 & .8854
 & 30\,\%\\
  \ac{farl} & .4410 ($\pm$.1435) & .0833 & .6406 & 20\,\%\\
  \midrule

\multicolumn{5}{l}{\textbf{Late Fusion}} \\
A + T & \textbf{\underline{.6128}} ($\pm$.1100) & .3806 & .8229 & 90\,\%\\
 A + V & .5493 ($\pm$.1079) & .3923 & .7604
 & 60\,\%\\
  T + V & .6101 ($\pm$.1074) & .4349 & .8281 & 90\,\%\\
  A + T + V & .6126 ($\pm$.1102) & .3794 & .8229 & 90\,\%\\
%   \midrule

\bottomrule
\end{tabular}
}
\caption{Results for direction prediction on the humorous segments, reported as \acp{AUC}. The columns are analogous to those of~\Cref{tab:humor_results}.}\label{tab:dir_results}
\end{table}

In~\Cref{tab:dir_results}, the results for the experiments on distinguishing humor direction in humorous segments are presented. Different from humor recognition, the visual modality performs worst, with all features only marginally above or even below chance on average. The standard deviations and maximum values, however, show that for a few coaches, facial features do encode information on whether humor is self- or others-directed. Similar phenomena can be observed in audio-based experiments. Here, more experiments yield results above chance, with \ds showing above-chance performance for 8 out of 10 coaches. On the one hand, the average audio results are only slightly larger than $.5000$ with no notable differences among the three different feature sets. On the other hand, there are coaches for which the audio modality alone is useful for predicting humor direction, as illustrated by the maximum \acp{AUC} for audio, \eg,  $.7656$ for \ds. 

The best mean \ac{AUC}s are obtained when experimenting with textual features. We hypothesize that the direction of an utterance is mainly encoded in its semantics and thus typically hard to predict from facial expressions or acoustic signals alone. Each textual feature outperforms all acoustic and visual features and the vast majority of textual experiments result in \acp{AUC} above chance level. Again, \textsc{ELECTRA} is the best performing text-based feature on average with a mean \ac{AUC} of $.6108$. The relative aptitude of text for predicting humor direction can be explained by direction being encoded in the semantics of an utterance. Oftentimes, it can be inferred from the content of the coach's comments whether he is joking about himself or others. Intuitively, this is typically not the case for facial expressions and paralinguistic features. However, the performance of textual features here probably suffers from the same problems already mentioned regarding textual humor recognition, namely partly highly domain-specific texts and lack of context due to not considering the interviewers' questions.

Overall, the direction is rather hard to predict, but at least the text modality yields some promising results. Moreover, results for direction prediction, too, are dependent on the subject, which is discussed in~\Cref{ssec:style_results}.

Because of the rather insufficient performance of audio and video features for direction prediction, they are not of much use in the late fusion experiments. Late fusion only slightly outperforms the best text-only experiment accounting for a mean \ac{AUC} of $.6128$ for the fusion of audio and text compared to $.6108$ for \textsc{ELECTRA} only. These findings suggest that, for direction prediction, multimodal approaches may not promise significant improvements over utilizing text only. They might even hurt the performance regarding those subjects that are rather hard to predict. For the trimodal fusion, the lowest \ac{AUC} value observed for one of the coaches is $.3794$, about $.0550$ lower than the \ac{AUC} of the worst performing coach in the \textsc{ELECTRA} experiments. Hence, it can be concluded that the direction of humor is mainly encoded in the semantics of an utterance.

\subsubsection{Sentiment}\label{sssec:sen_results}

\Cref{tab:sen_results} reports the results for sentiment prediction in humorous segments. In general, for each modality, higher \ac{AUC} values than for direction prediction (cf.~\Cref{tab:dir_results}) are achieved. However, this does not hold for all individual features, as the mean performances of \ds and \textsc{BERT} are slightly lower than in the direction experiments. 

\begin{table}[h!]
\resizebox{1\columnwidth}{!}{
\centering
\begin{tabular}{lrrrr}
\toprule
Feature & \multicolumn{1}{c}{Mean (Std)} & \multicolumn{1}{c}{Min} & \multicolumn{1}{c}{Max} & \multicolumn{1}{c}{$>$ Chance} \\ \midrule

\multicolumn{5}{l}{\textbf{Audio}} \\
\ac{egemaps} & \textbf{.5900} ($\pm$ .0579) & .5277 & .7109 & 100\,\%\\
 \ds & .5343 ($\pm$ .1085) & .2629 & .6613
 & 70\,\%\\
  \ac{W2V} & .5500 ($\pm$ .1285) & .1873 & .6845 & 90\,\%\\
  \midrule
  
  \multicolumn{5}{l}{\textbf{Text}} \\  
\textsc{BERT} & .5794 ($\pm$ .1165) & .2603 & .6932 & 90\,\%\\
 \textsc{GerSent-BERT} & .6141 ($\pm$ .0685) & .5175 & .7187 & 100\,\%\\
  \textsc{ELECTRA} & \textbf{.6446} ($\pm$ .0542) & .5512 & .7296  & 100\,\%\\
  \midrule
  
\multicolumn{5}{l}{\textbf{Video}} \\
\acp{FAU} & .6542 ($\pm$ .0749) & .5491 & .7884 & 100\,\%\\
 \vggf{} & .6342 ($\pm$ .0689) & .5127 & .7347 & 100\,\%\\
  \ac{farl} & \textbf{\underline{.6897}} ($\pm$ .0770) & .5531 & .8153 & 100\,\%\\

\midrule

\multicolumn{5}{l}{\textbf{Late Fusion}} \\
A + T & .6576 ($\pm$.0756) & .5920 & .7967 & 100\,\%\\
 A + V & .6482 ($\pm$.0732) & .5448 & .7732
 & 100\,\%\\
  T + V & \textbf{\underline{.7100}} ($\pm$.0748) & .6074 & .8294 & 100\,\%\\
  A + T + V & .6998 ($\pm$.0808) & .6023 & .8440 & 100\,\%\\
%   \midrule

\bottomrule
\end{tabular}
}\
\caption{Results for sentiment prediction on the humorous segments, reported as \acp{AUC}. The columns are analogous to those of~\Cref{tab:humor_results}.}\label{tab:sen_results}
\end{table}

Here, the \textbf{video} modality yields the best results on average, with \ac{farl} being responsible for the best mean \ac{AUC} value in total. The results for \textbf{text}, however, can compete with the visual ones, with \textsc{ELECTRA} as the best textual feature yielding $.6446$ mean \ac{AUC}, thus outperforming \vggf{} with $.6342$ mean \ac{AUC}. As expected by the construction of \textsc{GerSent-Bert} it outperforms the more generic \textsc{BERT} on the sentiment analysis task. Both visual and textual features almost always result in above-chance performance, the only exception being one coach for whom \textsc{BERT} fails with a mean \ac{AUC} of $.2603$. \textbf{Acoustic} features perform notably worse than visual and textual features on average, but at least the results for \ac{egemaps} always surpass chance level, while the \ac{W2V} features fail to do so only for one coach. Moreover, the mean \ac{AUC} of $.5900$ for \ac{egemaps} is higher than that of \textsc{BERT}. The relative superiority of textual features over acoustic ones is consistent with previous findings that the text modality is more suitable for valence prediction than acoustic data~\cite{calvo2010affect}.
Differences among coaches can be observed here, too, indicated by the standard deviations and notable differences between the minimum and maximum \ac{AUC} per coach for each feature, \eg, a difference of about $.2600$ for the \ac{farl} experiments.

In the late fusion experiments, there is evidence that the modalities partly complement each other. The combination of textual and visual predictions leads to a mean \ac{AUC} value of $.7100$, clearly outperforming the best visual-only ($.6897$) and textual-only ($.6446$) results. Moreover, the minimum \ac{AUC} encountered for a coach increases to $.6074$ in the fusion of video and text, while for both video and text only, the worst performing coach only achieves an \ac{AUC} of around $.5500$. While combining audio and video is thus beneficial, it seems like the audio modality could actually worsen the results. Fusing audio and video yields a lower mean \ac{AUC} value than the \ac{farl} experiments, with $.6482$ \ac{AUC} compared to $.6897$ and also the trimodal mean \ac{AUC} result is about $.0100$ lower than the result when only text and video are combined. We find that this phenomenon is due to massive overfitting to the training data when training \acp{SVM} using \ac{egemaps} features. On the training data, we find \acp{AUC} of over $.9000$, which causes audio-based predictions to receive larger weights than video and text-based predictions according to the algorithm described in~\Cref{sssec:late_fusion}, though the audio modality is, in comparison, the least suitable modality for sentiment prediction.

\subsection{Individual Results}\label{ssec:style_comparison}

Motivated by the partly large standard deviations and gaps between the best and worst performing coaches' \acp{AUC} observed in~\Crefrange{tab:humor_results}{tab:sen_results}, we analyze our results \wrt the individual coaches in the following.~\Cref{tab:ind_results} contains the best results obtained for every single coach in each of the three tasks and for each of the three modalities. Moreover, it provides each coach's rank per task based on the best result (across all modalities) for the respective coach.

\begin{table*}[h!]
% \resizebox{1\columnwidth}{!}{
\centering
\begin{tabular}{l|rrrr|rrrr|rrrr}
\toprule
 & \multicolumn{4}{c|}{\textbf{Humor}} & \multicolumn{4}{c|}{\textbf{Direction}} & \multicolumn{4}{c}{\textbf{Sentiment}} \\ 
    %   \cmidrule(lr){2-5} \cmidrule(lr){6-9} \cmidrule(lr){10-13}
 ID & \multicolumn{1}{c}{Audio} & \multicolumn{1}{c}{Text} & \multicolumn{1}{c}{Video} &\multicolumn{1}{c|}{Rank} & \multicolumn{1}{c}{Audio} & \multicolumn{1}{c}{Text} & \multicolumn{1}{c}{Video} &\multicolumn{1}{c|}{Rank}  & \multicolumn{1}{c}{Audio} & \multicolumn{1}{c}{Text} & \multicolumn{1}{c}{Video} &\multicolumn{1}{c}{Rank}  \\ \midrule \midrule

1 & .7763 & .7497 & \textbf{.9485} & 3 &
.6062 & \textbf{.7424} & .4994& 3 &
.6690   & .6932 & \textbf{\underline{.8153}}& 1
\\

2 & .8494 & .8827 & \textbf{\underline{.9625}}& 1  & 
\underline{.7656} & \underline{.8281} & \textbf{\underline{.8854}}& 1 &
.6222 & \textbf{.6984} & .6914& 6
\\

3 & .7682 & .8916 & \textbf{.9451}& 4  &
.5638 & \textbf{.7784} & .4843& 2   &
.6835 & \underline{.7296} & \textbf{.7833}& 2  
\\

4 & .7264 & .8118 & \textbf{.8899}& 8  &
.5342 & \textbf{.6039} & .4996& 10  &  
.5816 & .6527 & \textbf{.6902}& 7 
\\

5 & .6591 & .7703 & \textbf{.9193}& 6   &
.6153 & \textbf{.7040} & .6549& 5    &
.6613 & \textbf{.7187} & .6748& 5
\\

6 & .7528 & .7843 & \textbf{.9593}& 2   &
.6298 & \textbf{.6973} & .5069& 6    &
.5684 & \textbf{.6346} & .5531& 9 
\\

7 & .7089 & .8067 & \textbf{.8997}& 7   &
.5233 & \textbf{.6719} & .5193& 7 &
.6157 & .6466 & \textbf{.7473}& 4 
\\

8 & \underline{.8741} & \underline{.9169} & \textbf{.9383}& 5  & 
\textbf{.6688} & .6406 & .5420& 8   &
.5574 & .6764 & \textbf{.6868}& 8
\\

9 & .7036 & \textbf{.7266} & .7083& 10 &
\textbf{.6087} & .5352 & .5893& 9    &
.5441 & .5731 & \textbf{.6321}& 10
\\

10 & .7351 & .7442 & \textbf{.8255}& 9  &
.5331 & \textbf{.7122} & .4697& 4   &
\underline{.7109} & .7116 & .\textbf{7520}& 3
\\

\bottomrule
\end{tabular}
\caption{Comparison of results for individual coaches (denoted by \emph{ID}) for the tasks of humor recognition as well as direction and sentiment prediction. Reported are the best mean results per coach and modality. The best result per modality and task across all coaches is underlined. For every coach, the best result per task is boldfaced. \emph{R} denotes the rank of each coach for the three different tasks. The rank is determined by sorting the coaches according to their best mean \ac{AUC} value per task.}\label{tab:ind_results}
\end{table*}

Some coaches are generally easier to predict than others. Consider for example coaches 1 and 9. Coach 1 is ranked 3\textsuperscript{rd} for both humor recognition and direction prediction and first in sentiment prediction. In contrast, coach 9 ranks last in humor recognition and sentiment prediction and second-to-last in direction prediction, indicating that our approaches struggle with modeling his type of humor. There are also coaches whose humor can be detected relatively easily but this does not transfer to the classification of sentiment and direction. For example, for coach 6, an \ac{AUC} of $.9593$ is achieved for binary humor recognition, ranking him second in this task, while he only ranks 9\textsuperscript{th} and 6\textsuperscript{th} for sentiment and direction prediction, respectively. A similar phenomenon, yet reversed, can be observed for coach 10. While for humorous segments of this coach, both sentiment and direction are relatively easy to predict, ranking him 3\textsuperscript{rd} and 4\textsuperscript{th}, respectively, he ranks second-to-last in the binary humor detection task. 

% In order to get more quantitative insights into the relation between the rankings of the different tasks, we compute pearson correlations between the rankings. The pearson correlation between the humor and sentiment rankings is only $.1636$ while the correlation between the rankings for humor and dimension is $.6346$ and that between those for sentiment and direction is $.6485$. These findings suggest that, on average, 

% Rank correlations: h-s .1636, h-d: .6364, s-d: .6485 
% AUC correlations:
%   humor
%   A - T: 

For the recognition of humor,~\Cref{tab:humor_results} suggests that, for humor recognition, visual features outperform textual features which in turn lead to better results than audio features. This is confirmed by~\Cref{tab:ind_results} for most cases, except coaches 1 and 9. Coach 9 is, in general, hard to predict, and for humor, visual features do not work well for him, either. While for all other coaches, more than $.8000$ \ac{AUC} can be achieved, coach 9's humor \ac{AUC} with visual features is only $.7083$, inferior to his textual \ac{AUC} value of $.7266$. However, this coach was found to benefit the most from late fusion, improving his \ac{AUC} value to $.7825$ utilizing all three modalities. Audio outperforms text for humor recognition in only one case, namely coach 1.  

In the direction prediction task, the textual features work best for most of the coaches, consistent with the results discussed in~\Cref{sssec:dir_results}. However, similar to the observations for sentiment prediction, there also exist a few outliers. The highest \ac{AUC} values for coaches 8 and 9 are achieved using audio features, while for coach 9, the textual feature result of $.5352$ \ac{AUC} is only slightly above chance. For coach 1, the visual modality is responsible for the best result. Coach 1 and 5 are the only coaches, for whom \acp{AUC} of over $.6000$ can be observed when using visual features. 

As shown in~\Cref{tab:sen_results}, visual features typically yield higher \acp{AUC} than textual and audio features in the sentiment prediction task. In three cases, however, the text modality leads to better results than the video modality, most notably for coach 6, where both audio and text features outperform the \ac{AUC} value of $.5513$ obtained with visual features.
For the same coach, however, visual features perform well in humor detection, leading to the second-best humor \ac{AUC} among all coaches. Furthermore, it can be seen in~\Cref{tab:ind_results} that for some coaches all modalities achieve similar results, \eg, coach 10 in sentiment prediction, while for others, there are large discrepancies between the respective best \ac{AUC} values per modality, \eg, coach 1.  

To summarise, while there are general tendencies as to which modality works best for which of the three tasks, the results on the individual level often deviate from such general assumptions. This highlights the fact that humor styles differ among individuals and humor can be expressed in different ways. Hence, our models do not always generalize well for a previously unseen person.  

\revision{\section{Limitations}}
\revision{The results presented above shed more light onto the problem of humor detection, especially when attempting to model humor on the individual level. These insights, however, are limited by the peculiarities of our dataset.}
A significant shortcoming of \ac{Passau-SFCH} is that the set of subjects in the recordings is both small and homogeneous. All 10 coaches are male, share a similar cultural background, and work in the same profession. Moreover, their age span only ranges from 29-52\,years. A similar argument can be made regarding the demographically similar, non-random selection of annotators. Thus, many demographic groups are not represented in \ac{Passau-SFCH}, probably limiting the generalization abilities of models trained on it. 

Another limitation of our dataset is that though its setting is more spontaneous than in other existing datasets, the recordings are still, to a degree, staged. Press conferences are neither scripted nor truly `in-the-wild' interlocutions and professional football coaches can be expected to usually handle such communicative situations in a polite, matter-of-fact way. The professional context of press conferences may also impact the humor style displayed by the coaches. For example, it is likely that coaches use aggressive humor less frequently in such a situation than in private. Nevertheless, studies of similar Q\&A situations, such as earnings calls by top executives and securities analysts, highlight that Q\&A’s ``are relatively unscripted and spontaneous [so that executives, or sports coaches in our case,] are unable to fully rehearse or polish their answers''~\cite{graf2020effects}.

Furthermore, the recordings are highly domain-specific which may especially impair generalization in text-based approaches. The performance of text-based methods trained on \ac{Passau-SFCH} may also suffer from the fact that press conferences are actually multi-party conversations while in \ac{Passau-SFCH}, each video only comprises the data of one -- albeit the most important -- speaker per conference, \ie, the coach.  

While our dataset's homogeneity regarding domain, setting, and subjects' demographic features implies several shortcomings, it also accounts for a crucial advantage: individual variations in both the annotations and our experimental results (cf.~\Cref{ssec:style_results}) can be attributed to actual differences in the individual usage of humor rather than distortion caused by cultural or demographic differences. 

\section{Conclusion}\label{sec:conclusion}
We introduced \ac{Passau-SFCH}, a novel dataset for humor recognition that differs from existing datasets regarding the spontaneity and sparsity of the humorous utterances and its rich annotation scheme including the two dimensions of humor proposed in the \ac{HSQ}. 
Then, we investigated the suitability of different modalities (text, audio, and video) for the tasks of humor recognition as well as sentiment and direction classification. To do so, a rich set of features was employed. Our experiments show that humor in \ac{Passau-SFCH} is well-predictable, in particular when using the video modality. The sentiment of humorous segments proved harder to predict than humor, resulting in lower \ac{AUC} values than those observed for humor prediction. In sentiment prediction, the video modality, in general, led to the best results, too. The direction (self- vs other-directed) of humorous segments turned out to be the most difficult prediction target. Here, text stood out as the most promising modality. Moreover, for all three tasks, improvements via prediction-level fusion of models trained on different modalities could be noted. When analyzing the results on an individual level, we find substantial differences among the coaches in terms of \ac{AUC} values in general and for different modalities. 

These results strongly suggest that automatic humor recognition systems would benefit from personalization strategies. First experiments on personalizing \ac{ML} models trained on \ac{Passau-SFCH} have already been conducted~\cite{kathan22personalised}. Further future work may include developing more sophisticated multimodal approaches to address the three tasks proposed in this work. In particular, the whole conversation, not just the coaches' responses, could be taken into account. \revision{Another promising direction towards improving upon the baseline results is offered by recent pretrained multi-billion parameter text-only and multimodal \acp{LLM}}.
Another line of potential future work is the extension of the dataset to include additional subjects from more diverse backgrounds. In particular, this may entail collecting data from female coaches and coaches from different countries and cultures.

\revision{We make the \ac{Passau-SFCH} dataset available upon request for academic non-commercial research purposes\footnote{\href{https://www.github.com/lc0197/passau-sfch}{https://www.github.com/lc0197/passau-sfch}}}.
% % Appendix two text goes here.

\ifCLASSOPTIONcompsoc
  \section*{Acknowledgments}
\else
  \section*{Acknowledgment}
\fi

This research was supported by Deutsche Forschungsgemeinschaft (DFG) under grant agreement No.\ 461420398 %BS: Grantnummer alleine hat keinen Wiedererkennungseffekt - ich habe eingefügt (bitte prüfen!):
(``Leader Humor''). Shahin Amiriparian and Bj\"orn W. Schuller are also with MDSI -- Munich Data Science Institute as well as MCML -- Munich Center of Machine Learning. Bj\"orn W. Schuller is also with the Konrad Zuse School of Excellence in Reliable AI (relAI), Munich, Germany.

\ifCLASSOPTIONcaptionsoff
  \newpage
\fi

\newpage



\footnotesize
\bibliographystyle{IEEEtranN}
% \bibliography{IEEEabrv,../bib/paper}
\bibliography{refs}



% \end{thebibliography}

% Maurice
% \begin{IEEEbiography}[{\includegraphics[width=1in,height=1.25in,clip,keepaspectratio]{photos/maurice}}]{Maurice Gerczuk}
% received his M.Sc. in Computer Science at the University of Passau. There, he worked with the chair of Complex and Intelligent Systems as student research assistant. Currently, he is a doctoral student at the chair of Embedded Intelligence for Health-Care and Wellbeing at the University of Augsburg, Germany. His research focuses on intelligent signal processing for health care and affective computing. He works on ParaStiChaD.
% \end{IEEEbiography}	
% \vspace{-1cm}

\begin{acronym}
\acro{A}[A]{Arousal}
\acro{ABC}[ABC]{Airplane behavior Corpus}
\acro{AD}[AD]{Anger Detection}
\acro{AFEW}[AFEW]{Acted Facial Expression in the Wild)}
\acro{AI}[AI]{Artificial Intelligence}
\acro{ANN}[ANN]{Artificial Neural Network}
\acro{ASO}[ASO]{Almost Stochastic Order}
\acro{ASR}[ASR]{Automatic Speech Recognition}
\acro{AUC}[AUC]{Area Under the Curve}

\acro{BBFN}[BBFN]{Bi-bimodal Fusion Network}
\acro{BN}[BN]{batch normalisation}
\acro{BiLSTM}[BiLSTM]{Bidirectional Long Short-Term Memory}
\acro{BFI-S}[BFI-S]{Big Five Inventory-SOEP}
\acro{BES}[BES]{Burmese Emotional Speech}
\acro{BoAW}[BoAW]{Bag-of-Audio-Words}
\acro{BoDF}[BoDF]{Bag-of-Deep-Feature}
\acro{BoW}[BoW]{Bag-of-Words}

\acro{CASIA}[CASIA]{Speech Emotion Database of the Institute of Automation of the Chinese Academy of Sciences}
\acro{CCC}[CCC]{Concordance Correlation Coefficient}
\acro{CVE}[CVE]{Chinese Vocal Emotions}
\acro{CNN}[CNN]{Con\-vo\-lu\-tion\-al Neural Network}
\acro{ComParE}[ComParE]{Computational Paralinguistics Challenge}
\acrodefplural{ComParE}[ComParE]{Computational Paralinguistics Challenges}
\acro{CRF}[CRF]{Conditional Random Field}
\acrodefplural{CRF}[CRFs]{Conditional Random Fields}

\acro{CRNN}[CRNN]{Con\-vo\-lu\-tion\-al Recurrent Neural Network}

\acro{DEMoS}[DEMoS]{Database of Elicited Mood in Speech}
\acro{DES}[DES]{Danish Emotional Speech}
\acro{DNN}[DNN]{Deep Neural Network}
\acro{DS}[DS]{\ds}

\acro{egemaps}[eGeMAPS]{extended version of the Geneva Minimalistic Acoustic Parameter Set}
\acro{EMO-DB}[EMO-DB]{Berlin Database of Emotional Speech}
\acro{EmotiW}[EmotiW 2014]{Emotion in the Wild 2014}
\acro{EWE}[EWE]{Evaluator Weighted Estimator}
\acro{eNTERFACE}[eNTERFACE]{eNTERFACE'05 Audio-Visual Emotion Database}
\acro{EU-EmoSS}[EU-EmoSS]{EU Emotion Stimulus Set}
\acro{EU-EV}[EU-EV]{EU-Emotion Voice Database}

\acro{farl}[FaRL]{FaRL for Facial Representation Learning}
\acro{FAU}[FAU]{Facial Action Unit}
\acrodefplural{FAU}[FAUs]{Facial Action Units}

\acro{AIBO}[FAU Aibo]{FAU Aibo Emotion Corpus}
\acro{FCN}[FCN]{Fully Convolutional Network}
\acro{FFT}[FFT]{fast Fourier transform}

\acro{GAN}[GAN]{Generative Adversarial Network}
\acro{GEMEP}[GEMEP]{Geneva Multimodal Emotion Portrayal}
\acro{GMU}[GMU]{Gated Multimodal Unit}
\acro{GRU}[GRU]{Gated Recurrent Unit}
\acro{GVEESS}[GVEESS]{Geneva Vocal Emotion Expression Stimulus Set}

\acro{HSQ}[HSQ]{Humor Style Questionnaire}

\acro{IEMOCAP}[IEMOCAP]{Interactive Emotional Dyadic Motion Capture}
\acro{IoU}[IoU]{Intersection over Union}

\acro{LSTM}[LSTM]{Long Short-Term Memory}
\acro{LLD}[LLD]{low-level descriptor}
\acro{LLM}[LLM]{Large Language Model}
\acro{LMF}[LMF]{Low-rank Multimodal Fusion}

\acro{ML}[ML]{Machine Learning}
\acro{MELD}[MELD]{Multimodal EmotionLines Dataset}
\acro{MES}[MES]{Mandarin Emotional Speech}
\acro{MFCC}[MFCC]{Mel-Frequency Cepstral Coefficient}
\acro{MIP}[MIP]{Mood Induction Procedure}
\acro{MISA}[MISA]{Modality-Invariant and -Specific Representations for Multimodal Analysis}
\acro{MLP}[MLP]{Multilayer Perceptron}
\acro{MuSe}[MuSe]{\textbf{Mu}ltimodal \textbf{Se}ntiment Analysis Challenge}
\acro{MuSe-Humor}[\textsc{MuSe-Humor}]{Humor Detection Sub-Challenge}
\acro{MulT}[MulT]{Multimodal Transformer}
\acro{NLP}[NLP]{Natural Language Processing}
\acro{NLU}[NLU]{Natural Language Understanding}

\acro{Passau-SFCH}[\textsc{Passau-SFCH}]{Passau Spontaneous Football Coach Humor}

\acro{ReLU}[ReLU]{Rectified Linear Unit}
\acro{RMSE}[RMSE]{root mean square error}
\acro{RNN}[RNN]{Recurrent Neural Network}
\acrodefplural{RNN}[RNNs]{Recurrent Neural Networks}

\acro{SER}[SER]{Speech Emotion Recognition}
\acro{SGD}[SGD]{Stochastic Gradient Descent}
\acro{SVM}[SVM]{Support Vector Machine}
\acro{SIMIS}[SIMIS]{Speech in Minimal Invasive Surgery}
\acro{SmartKom}[SmartKom]{SmartKom Multimodal Corpus}
\acro{SUSAS}[SUSAS]{Speech Under Simulated and Actual Stress}

\acro{UAR}[UAR]{Unweighted Average Recall}
\acro{V}[V]{Valence}
\acro{VFMM}[VFMM]{Video-Focused Multimodal Transformer}
\acro{WSJ}[WSJ]{Wall Street Journal}
\acro{W2V}[W2V]{Wav2Vec2}
\end{acronym}

\newpage

\begin{IEEEbiography}[{\includegraphics[width=1in,height=1.25in,clip,keepaspectratio]{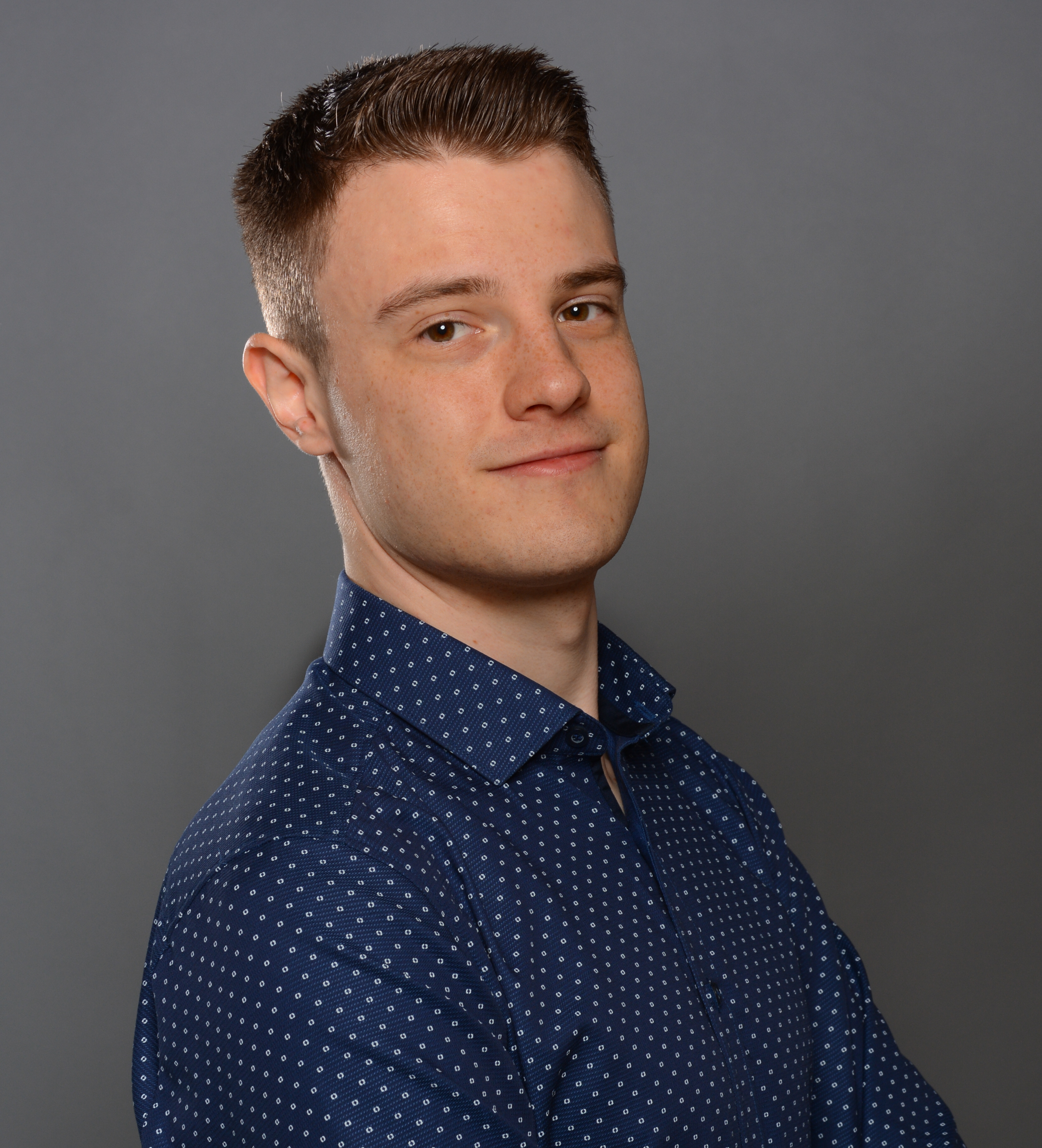}}]{Lukas Christ}
 received his Master's degree in Computer Science at the University of Leipzig in 2020. He is currently a PhD candidate at the Chair of Embedded Intelligence for Health Care and Wellbeing at the University of Augsburg, Germany. His main research interests are natural language processing and multimodal machine learning in the context of affective computing. 
%He was a co-organizer of the Multimodal Sentiment Analysis Challenge and Workshop (MuSe) at ACM MM 2022.
\end{IEEEbiography}	

\begin{IEEEbiography}[{\includegraphics[width=1in,height=1.25in,clip,keepaspectratio]{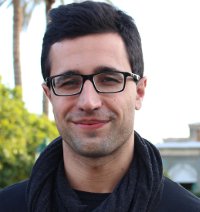}}]{Shahin Amiriparian}
 received his Doctoral degree with the highest honors (summa cum laude) at the Technical University of Munich (TUM), Germany, in 2019. He is an assistant professor at the Chair of Health Informatics, Klinikum rechts der Isar, TUM, Germany. His main research focus is digital health, affective computing, and multimodal signal processing. He (co-)authored more than 100 publications in peer-reviewed books, journals, and conference proceedings (h-index: 26) and is involved in various international conferences as program chair and workshop organizer.
\end{IEEEbiography}

\begin{IEEEbiography}[{\includegraphics[width=1in,height=1.25in,clip,keepaspectratio]{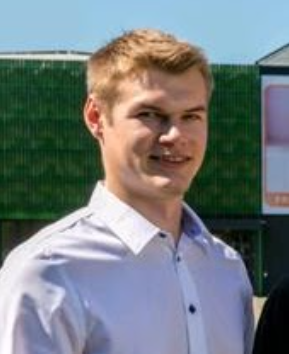}}]{Alexander Kathan}
 received his M.\,Sc.\ degree in Business Analytics from the University of Ulm, Germany, in 2021. Currently, he is pursuing his Ph.\,D.\ degree with the Chair of Embedded Intelligence for Health Care and Wellbeing at the University of Augsburg, Germany. His research interests include deep learning and machine learning methods for audio and multimodal signal processing in healthcare applications, as well as personalised machine learning approaches.
\end{IEEEbiography}

% Sandra
% \begin{IEEEbiography}[{\includegraphics[width=1in,height=1.25in,clip,keepaspectratio]{photos/sandra}}]{Sandra Ottl}
% received her M.Sc. in Computer Science at the University of Passau. There, she worked at the chair of Complex and Intelligent Systems as student research assistant. Currently, she is pursuing her doctoral degree as a researcher at the chair of Embedded Intelligence for Health Care and Wellbeing at the University of Augsburg, Germany and works on the KIRUN project.
% \end{IEEEbiography}	
% \vspace{-1cm}

% Niklas
\begin{IEEEbiography}[{\includegraphics[width=1in,height=1.25in,clip,keepaspectratio]{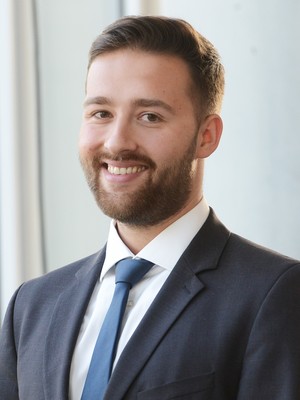}}]{Niklas M\"uller}
 received his M.\,A.\ HSG degree in Strategy and International Management (SIM) from the University of St.\ Gallen, Switzerland in 2019. Currently, he is pursuing his Doctoral Degree with the Chair of Strategic Management, Innovation, and Entrepreneurship at the University of Passau, Germany. His research interests are strategic communication of leaders and executives, in particular humor.
\end{IEEEbiography}	

% Andi
\begin{IEEEbiography}[{\includegraphics[width=1in,height=1.25in,clip,keepaspectratio]{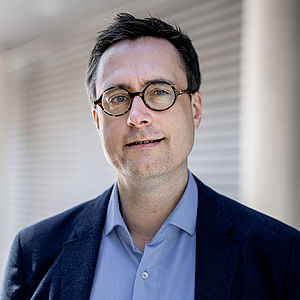}}]{Andreas K\"onig} received his Master of Music from the Royal Academy of Music, London, an MBA at HHL Leipzig, his Doctoral Degree at FAU Erlangen-Nürnberg. Since 2013, he holds the chair of Strategic Management, Innovation, and Entrepreneurship at the University of Passau, Germany. He is also the Deputy Director of the DFG Research Training Group “Digital Platform Ecosystems,” and he has been visiting professor at, for example, Free University Amsterdam and WU Vienna. His research focuses on topics at the interface of strategic leadership, discontinuous change, and executive communication and has appeared in leading outlets, such as Administrative Science Quarterly, Academy of Management Review, and Academy of Management Journal.
\end{IEEEbiography}	

% Bjoern
\begin{IEEEbiography}[{\includegraphics[width=1in,height=1.25in,clip,keepaspectratio]{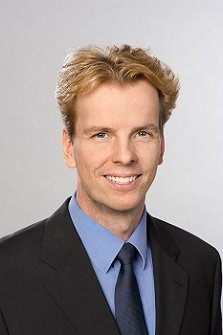}}]{Bj\"orn W. Schuller} is Professor of Artificial Intelligence at the Imperial College London, UK, where he heads GLAM -- the Group on Language, Audio, and Music, Full Professor and Chair of Health Informatics at TUM in Munich, Germany, Visiting Professor at HIT, China, and founding CEO and current CSO of audEERING, Germany known for audio-based sentiment and emotion analysis. Prof.\ Schuller (co-)authored more than 1,500 publications leading to $>$60,000 citations (h-index: 111). He is Field Chief Editor of the Frontiers in Digital Health, Editor in Chief of the AI Open Journal, and was Editor in Chief of the IEEE Transactions on Affective Computing, and served as General Chair for ACII 2019, ACII Asia 2018, and ICMI 2014, Program Chair of Interspeech 2019, ACM ICMI 2019, 2013, ACII 2015, 2011, and IEEE SocialCom 2012, and initiated and co-organised $>$40 international challenges, including the annual ComParE challenge series since 2009, and the annual Audio/Visual Emotion Challenge and Workshop (AVEC) series from 2011 to 2019. He is President-Emeritus and Fellow of the AAAC, Fellow of the ACM, Fellow of the BCS, Fellow of the ELLIS, Fellow of the IEEE, and Fellow of the ISCA.
\end{IEEEbiography}	

\newpage

\newpage

\end{document}